\documentclass[journal]{new-aiaa}
\usepackage[utf8]{inputenc}
\usepackage{textcomp}
\usepackage{graphicx}
\usepackage{amsmath,amsfonts}
\usepackage[version=4]{mhchem}
\usepackage{siunitx}
\usepackage{longtable,tabularx}
\usepackage[dvipsnames]{xcolor}
\usepackage{lipsum}  
\usepackage{multicol}
\usepackage{tikz}
\usepackage{float}
\usepackage{bm}
\usepackage{mathtools}
\usepackage{algorithm}
\usepackage{algpseudocodex}
\usetikzlibrary{shapes}
\usetikzlibrary{decorations.pathreplacing}
\usetikzlibrary{fadings}
\newcommand{\pr}[1]{\left(#1\right)}
\newcommand{\sq}[1]{\left[#1\right]}
\newcommand{\br}[1]{\left\{#1\right\}}

\title{Precision Mars Entry Navigation with Atmospheric Density Adaptation via Neural Networks}

\author{Felipe Giraldo-Grueso\footnote{Ph.D. Student, Department of Aerospace Engineering and Engineering Mechanics; fgiraldo@utexas.edu} and Andrey A. Popov\footnote{Postdoctoral Fellow, Oden Institute for Computational Engineering and Sciences; 
andrey.a.popov@utexas.edu.} and Renato Zanetti\footnote{Associate Professor, Department of Aerospace Engineering and Engineering Mechanics, AIAA Associate Fellow; renato@utexas.edu}}
\affil{The University of Texas at Austin, Austin, TX 78712}
\begin{document}

\maketitle

\begin{abstract}
Spacecraft entering Mars require precise navigation algorithms capable of accurately estimating the vehicle's position and velocity in dynamic and uncertain atmospheric environments. Discrepancies between the true Martian atmospheric density and the onboard density model can significantly impair the performance of spacecraft entry navigation filters. This work introduces a new approach to online filtering for Martian entry using a neural network to estimate atmospheric density and employing a consider analysis to account for the uncertainty in the estimate. The network is trained on an exponential atmospheric density model, and its parameters are dynamically adapted in real time to account for any mismatch between the true and estimated densities. The adaptation of the network is formulated as a maximum likelihood problem by leveraging the measurement innovations of the filter to identify optimal network parameters. Within the context of the maximum likelihood approach, incorporating a neural network enables the use of stochastic optimizers known for their efficiency in the machine learning domain. Performance comparisons are conducted against two online adaptive approaches, covariance matching and state augmentation and correction, in various realistic Martian entry navigation scenarios. The results show superior estimation accuracy compared to other approaches, and precise alignment of the estimated density with a broad selection of realistic Martian atmospheres sampled from perturbed Mars-GRAM data.
\end{abstract}

\section{Introduction}

\lettrine{A}{mong} the most demanding challenges encountered in spacecraft navigation is atmospheric entry on Mars. The entry phase is characterized by intense dynamics, a scarcity of available measurements, and uncertain atmospheric information~\cite{zanetti2007}. Due to the uncertainty in atmospheric conditions, discrepancies between the true atmospheric density and onboard models can make navigation during entry challenging and significantly degrade guidance solutions, affecting landing accuracy~\cite{tracy2023}. Despite improvements in the knowledge of the Martian atmosphere, its high variability, coupled with a scarcity of comprehensive data, makes it less accurately predictable compared to Earth's atmosphere~\cite{zanetti2007}, complicating the development of standard Entry, Descent, and Landing (EDL) systems. Standard EDL systems typically use a correction mechanism that adjusts an exponential density profile with a multiplicative factor based on the drag force experienced by the spacecraft~\cite{putnam2014}. However, constraining density estimation to an exponential profile can pose challenges for Martian entry. For example, the significant atmospheric dust content on Mars is known to increase the temperature of the lower atmosphere and subsequently reduce its density~\cite{liu2018}. As a result, future human and robotic Mars exploration missions require improvements by several orders of magnitude over the current state-of-the-art~\cite{braun2006}.

Previous Mars entry missions have used inertial measurement units (IMU) for navigation instead of onboard models of non-conservative forces. However, IMU-based solutions tend to drift over time, especially without external measurements. If standalone IMU measurements are used for state correction rather than propagation, most estimation states can become unobservable during entry~\cite{levesque2006}. Therefore, various techniques have been explored to improve navigation solutions during Martian entry. Zanetti and Bishop~\cite{zanetti2007b} successfully quantified the accuracy of state estimates through dead-reckoning during standard Martian atmospheric entry only using IMU measurements for navigation. Following this approach, Lou et al.~\cite{lou2015} presented a strategy attempting to mitigate the effects of unobservable uncertain parameters through consider analysis during entry. Additionally, Jiang et al.~\cite{jiang2018} introduced a more accurate multisensor strategy, where instead of using only an IMU for navigation, information from a real-time flush air data system, a single orbiting radio beacon, and an IMU were fused using an unscented Kalman filter, providing better estimation results. Recently, the integration of aerodynamic pressure and aerothermal sensor data for real-time navigation has been proposed~\cite{zegwe2023}. However, most of these solutions face a common challenge: they either rely on an onboard exponential atmospheric density model or depend on external measurements, such as orbiting or surface beacons. The assumption of an exponential density model may pose difficulties when faced with real density profiles, potentially leading to filter divergence. Furthermore, due to the presence of a heat-shield cover on the entry vehicle, onboard optical or radio navigation may not be available during entry~\cite{zegwe2023}, making external measurements unavailable.

The descent and landing phases are typically supported by sensors, such as altimeters, velocimeters, or LIDARs, which can aid navigation solutions~\cite{marcus2022,marcus2023}. However, these sensors are not available during the entry phase, which poses the most significant challenge for navigation. Accurate density estimation becomes crucial in the absence of additional sensors. Therefore, effectively estimating atmospheric density and quantifying its uncertainty can significantly enhance navigation outputs, improving precision landing performance. A range of methodologies have been investigated to address atmospheric uncertainties within the Martian environment with the aim of improving the performance of EDL systems. These approaches can be categorized into two main groups: offline reconstructions, which involve working with real data acquired during the entry phase, and online estimation techniques, which are designed for use during the entry phase. Dutta et al.~\cite{dutta2013} analyzed various estimation techniques, including the extended Kalman filter (EKF) and unscented Kalman filter (UKF) for offline reconstruction. In their work, the main objective was to reconstruct the vehicle trajectory, aerodynamic coefficients, and atmospheric profile experienced by the system in an offline context. Subsequently, in their follow-up work~\cite{dutta2014}, they demonstrated how adaptive filtering techniques such as covariance matching~\cite{mehra1972,myers1976} could enhance the estimation performance of aerodynamic coefficients and atmospheric properties. These advanced techniques were later employed to successfully reconstruct atmospheric properties using real data acquired by the Mars Science Laboratory (MSL)~\cite{dutta2014b}.

To improve filtering precision in online estimation for EDL, various strategies have been explored. A common solution involves carrying a nominal density onboard the filter and augmenting the system state to estimate a correction factor that accounts for the deviation between the nominal density and actual density experienced by the system~\cite{zegwe2023,tracy2023}. In addition, machine learning approaches have been used for density estimation in EDL problems. Wagner et al.~\cite{wagner2011} used an ensemble of neural networks to dynamically adapt to real-time changes in atmospheric properties during Martian entry. Additionally, Amato and McMahon~\cite{amato2021} employed a long short-term memory network to estimate atmospheric density based on inertial measurements and guidance commands. Most recently, Roelke et al.~\cite{roelke2023} introduced two density profile prediction techniques for aerocapture, using density histories computed from navigation measurements to enhance targeting accuracy.

In this work, we propose a new approach to onboard adaptive filtering for Martian entry by using a neural network, trained pre-flight on a simplified model retrieved from Mars-GRAM trajectories, to estimate atmospheric density. To address the atmospheric mismatch between the true density and the neural network, the network's parameters are adapted in real time using measurement innovations. The measurement innovations are defined as the difference between the true measurements and the predicted measurements of the navigation filter. Furthermore, we account for the uncertainty in the atmospheric density estimate using consider analysis. Consider analysis acknowledges errors in the dynamic and measurement models due to uncertain parameters~\cite{zanetti2012}, such as atmospheric density in this case. This approach enables the navigation filter to correctly weight the data while accounting for the correlation between the state and the uncertain parameter~\cite{stauch2015}. The adaptation of the network is performed using aerodynamic pressure measurements, aerothermal sensor data, and an IMU, mirroring sensor configurations from prior Martian EDL missions~\cite{gazarik2008,hwang2016,white2022}. Leveraging the capabilities of state-of-the-art stochastic optimizers, nonlinear activation functions, and automatic differentiation, the adaptation step is posed as a maximum likelihood (ML) problem, resulting in consistent filtering performance and accurate estimation of the true density.

The remainder of this paper is organized as follows: First, a background of Martian entry is provided in Section~\ref{sec:marsentry}. Previous approaches to density adaptation are introduced in Section~\ref{sec:densityadap}, and the approach to density adaptation using neural networks is presented in Section~\ref{sec:densityadaptwnn}. The performance of the new approach is compared to previous approaches on a realistic Mars entry navigation scenario in Section~\ref{sec:resultsanddiscussion}. Section~\ref{sec:limitations} discusses some of the limitations of the framework presented and outlines future work opportunities. Finally, Section~\ref{sec:conclusions} presents the conclusions of the work.

\section{Martian Entry}\label{sec:marsentry}

This section provides insight into the dynamics, measurement, and atmospheric density models used in Martian entry. The aim of this analysis is to emphasize the importance of acquiring an accurate atmospheric density estimate during the entry phase.

\subsection{Dynamics}\label{sec:dynamics}

\begin{figure}
    \centering
    \includegraphics[width = 0.5\textwidth]{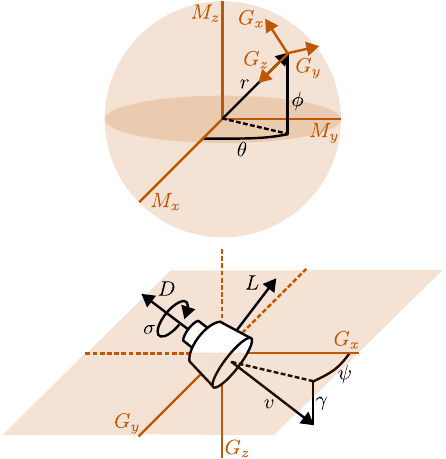}
    \caption{Position coordinates with respect to the MCMF frame $M$ (top) and velocity coordinates with respect to the Geographic frame $G$ (bottom).}
    \label{fig:angles}
\end{figure}

The dynamics of entry vehicles during Martian entry are often represented using a simplified three degrees-of-freedom (3-DOF) model. In this approach and for the entirety of this work, point mass dynamics are assumed, meaning that no consideration is given to attitude dynamics, a common convention observed in prior literature~\cite{vinh1980,levesque2006,christian2007,lu2014,zegwe2023}. The position of the vehicle is described in terms of planet-centric radius ($r$), latitude ($\phi$), and longitude ($\theta$) which are shown with respect to the Mars-centered Mars-fixed (MCMF) frame~\cite{levesque2006} in Fig.~\ref{fig:angles}. The MCMF frame is centered and fixed to the planet aligning $M_z$ with the axis of rotation. These coordinates allow the use of spherical equations of motion, even for non-spherical planets, as the radius can vary with latitude and longitude~\cite{christian2007}. The dynamic equations used to model the evolution of these variables with respect to time are given by ~\cite{levesque2006,lu2014,zegwe2023},
\begin{align}
    \dot{r} & = v\sin(\gamma),\\
    \dot{\phi} & =\frac{\cos(\gamma)\cos(\psi)}{r}{v}, \\
    \dot{\theta} & = \frac{\cos(\gamma)\sin(\psi)}{r\cos(\phi)} v.
\end{align}

The magnitude of the velocity vector ($v$), flight path angle ($\gamma$), and heading azimuth ($\psi$) are used to describe the velocity of the vehicle as shown with respect to the Geographic frame~\cite{levesque2006} in Fig.~\ref{fig:angles}. The Geographic frame is centered on the vehicle and $G_x$, $G_Y$, $G_z$ are aligned with Mars geographic directions North, East and Nadir, respectively. The flight path angle is defined as negative-down, and the heading azimuth is defined in the horizontal plane with $0^\circ$ as due North and $90^\circ$ as due East. The equations describing the time evolution of the velocity-related terms are given by~\cite{levesque2006,lu2014,zegwe2023}:
\begin{align}
    \dot{v} & = -D - g\sin(\gamma) + \omega^2r\cos(\phi) \left[\sin(\gamma)\cos(\phi)  - \cos(\gamma)\sin(\phi)\cos(\psi)\right], \\
    \dot{\gamma} & = \frac{1}{v}\left[L\cos(\sigma) - g\cos(\gamma) + \frac{v^2}{r}\cos(\gamma) + 2\omega v \cos(\phi)\sin(\psi)\right] \\ & + \frac{1}{v}\sq{\omega^2 r \cos(\phi)\left(\cos(\gamma)\cos(\phi) + \sin(\gamma)\sin(\phi)\cos(\psi)\right)}\nonumber,\\
    \dot{\psi} & = \frac{1}{v}\sq{L\frac{\sin(\sigma)}{\cos(\gamma)}+\frac{v^2}{r}\cos(\gamma)\sin(\psi)\tan(\phi) - 2\omega v\pr{\tan(\gamma)\cos(\phi)\cos(\psi) - \sin(\phi)}}\\ & + \frac{1}{v}\sq{\frac{\omega^2r}{\cos(\gamma)}\sin(\phi)\cos(\phi)\sin(\psi)}\nonumber,
\end{align}
where $\omega$ represents the planetary rotation rate, $g$ is the planetary gravity, $\sigma$ is the bank angle, and the lift ($L$) and drag ($D$) accelerations are defined as follows~\cite{levesque2006}:
\begin{align}
    D & = \frac{1}{2}\rho\pr{\cdot}v_{\infty}^2 \frac{C_d S}{m}, \\ L  & = \frac{1}{2}\rho\pr{\cdot} v_{\infty}^2 \frac{C_l S}{m},
\end{align}
with $v_\infty = v - W$, where $W$ represents the wind velocity. The parameter $\rho\pr{\cdot}$ is the atmospheric density, where the notation $(\cdot)$ signifies its dynamic nature during the entry phase. Given the dependency of lift and drag accelerations on atmospheric density, all velocity-related terms become directly influenced by changes in atmospheric density. As a result, the temporal evolution of the position implicitly relies on atmospheric density variations.

Given the substantial hypersonic speeds associated with the entry phase, wind velocities are considerably lower compared to the velocity of the vehicle, i.e., $v/W\ll 1$~\cite{amato2021}. In such cases, it is reasonable to make the approximation that $v_\infty \approx v$. Moreover, for brief-duration flights such as the entry phase in EDL, the impact of planetary rotation can be deemed negligible~\cite{zegwe2023}. Note that neglecting planetary rotation makes frame $M$ a simple inertial frame. Hence, by omitting terms related to planetary rotation and assuming a first-order gravity model, the equations governing velocity parameters simplify to~\cite{levesque2007}:
\begin{align}
    \dot{v} & = -\frac{1}{2}\rho\pr{\cdot} v^2 \frac{C_d S}{m} -\frac{\mu}{r^2}\sin(\gamma), \\
    \dot{\gamma} & = \frac{1}{v}\left[\frac{1}{2}\rho\pr{\cdot} v^2 \frac{C_l S}{m}\cos(\sigma) -\frac{\mu}{r^2}\cos(\gamma) + \frac{v^2}{r}\cos(\gamma) \right], \\
    \dot{\psi} & = \frac{1}{v}\sq{\frac{1}{2}\rho\pr{\cdot} v^2 \frac{C_l S}{m}\frac{\sin(\sigma)}{\cos(\gamma)}+\frac{v^2}{r}\cos(\gamma)\sin(\psi)\tan(\phi)}.
\end{align}

By reviewing the dynamics that govern Martian entry, it becomes evident that errors in the estimation of atmospheric density can significantly impact the dynamics, leading to a considerable degradation in the propagation of filtering states. This emphasizes just how crucial it is to attain an accurate estimate of density for navigation solutions.

\subsection{Onboard Sensors}\label{sec:measurements}

The measurement model of the onboard sensors typically used in EDL missions is closely tied to the atmospheric density estimate as well. In this work, three distinct onboard sensors are under consideration. These sensors encompass an IMU, a cluster of pressure sensors strategically placed on the heat shield of the vehicle, and a suite of thermocouples situated on the forebody of the vehicle. These sensors are carefully selected to replicate the sensor configurations used in past Martian EDL missions~\cite{zegwe2023,gazarik2008,hwang2016,white2022}. 

\subsubsection{Inertial measurement unit}

The IMU measures the non-gravitational accelerations encountered by the entry vehicle. In this work, these measurements are conducted relative to the body frame aligned with the axes of symmetry in the vehicle~\cite{zegwe2023}. The acceleration measurements are described by:
\begin{equation}
    \Tilde{\bm{a}}^b = T_v^b\bm{a}^v + \bm{\eta}_{\bm{a}},
\end{equation}
where $\bm{\eta}_{\bm{a}} \in \mathbb{R}^3$ represents measurement noise, $T_v^b$ is the transformation matrix to change coordinates from velocity frame to body frame, and $\bm{a}^v \in \mathbb{R}^3$ are the non-gravitational accelerations in the velocity frame defined as~\cite{zegwe2023}:
\begin{equation}
    \bm{a}^v = \begin{bmatrix} -D & L\sin\pr{\sigma} & L \cos\pr{\sigma}\end{bmatrix}^\mathrm{T}.
\end{equation}

The velocity frame is centered on the vehicle, with its $x$-axis aligned in the direction of the flight path angle, its $z$-axis aligned with the local vertical plane orthogonal to the $x$-axis, and the $y$-axis completing the right-handed system~\cite{zegwe2023}. 

\subsubsection{Cluster of pressure sensors}

The pressure sensors are assumed to offer a reasonably accurate evaluation of the stagnation-point pressure. It is important to acknowledge that the precision of these measurements may vary according to the flight regime. For instance, pressure transducers optimized for hypersonic flight may not deliver satisfactory results during supersonic flight. Nevertheless, this work assumes the availability of measurements throughout the entire entry stage. The dynamic pressure measurement, assuming an aggregate measurement from all relevant sensor arrays, is expressed as follows~\cite{zegwe2023}:
\begin{equation}
    \tilde{q} = \frac{1}{2}\rho\pr{\cdot} v^2 + \eta_q,
\end{equation}
where $\eta_{q}\in \mathbb{R}$ represents measurement noise. 

\subsubsection{Suite of thermocouples}

The suite of thermocouples is assumed to provide real-time measurements of the convective heating rate. The measured convective heating rate, assuming again an aggregate measurement from all relevant sensor arrays, is modeled by the Sutton-Graves relation~\cite{sutton1971,benito2010},
\begin{equation}
    \tilde{\dot{Q}}_s = k\pr{\frac{\rho\pr{\cdot}}{R_n}}^{\frac{1}{2}}v^3 + \eta_{\dot{Q}_s},
\end{equation}
where $k = 1.9027 \times 10^{-4}$ kg$^{1/2}$m$^{-1}$, $R_n$ is the vehicle nose radius and $\eta_{\dot{Q}_s}\in \mathbb{R}$ is measurement noise. 

These equations further emphasize the critical importance of having an accurate atmospheric density estimate, as they all rely on this fundamental parameter. This connection illustrates how easily a navigation system can diverge when confronted with inaccurate estimates or inaccurate density models stored in the onboard computer of the vehicle, emphasizing the role of obtaining a highly accurate density estimate for mission success and safety.

\subsection{Density Models}

To model Martian atmospheric density, various approaches are available, ranging from more intricate models for increased accuracy to simpler equations for faster estimates, although with reduced precision. 

\subsubsection{Exponential models}

The most straightforward equation for modeling atmospheric density is based on an exponential law, assuming isothermal equilibrium~\cite{regan1993}. This model is represented by the equation:
\begin{equation}\label{eq:exponential}
\hat{\rho}_{\exp} = \rho_0\exp\left(-\frac{(r - r_0)}{h_s}\right),
\end{equation}
where $r_0$ represents the reference radial position, $\rho_0$ is the density at the reference radial position, and $h_s$ stands for the atmospheric scale height. It is worth noting that while the atmosphere is not isothermal at all altitudes, the parameters in \eqref{eq:exponential} can be adjusted piece-wise to better align with the real atmospheric density. However, as previously mentioned, constraining Martian density to an exponential profile can be challenging due to the variability in dust content, seasonal changes, and wind patterns~\cite{liu2018}. In response to potential errors in the exponential model, the COSPAR model was developed using data collected from Viking 1, Viking 2, and Mariner missions~\cite{zanetti2007}. Although lacking consideration for factors like longitude, latitude, seasonal variations, and dust content, the COSPAR model offers more precision than the basic exponential model. The simplified COSPAR model is expressed as a modified exponential equation,
\begin{equation}
    \hat{\rho}_{\mathrm{COSPAR}} = \rho_0\exp\br{-\beta_\rho\pr{r - r_0} + \gamma_\rho\cos\sq{\omega_\rho \pr{r - r_0}} + \delta_\rho\sin\sq{\omega_\rho \pr{r - r_0}}},
\end{equation}
where $\beta_\rho, \gamma_\rho, \omega_\rho$ and $\delta_\rho$ are nominal coefficients adjustable to better fit real atmospheric data. Both the exponential and COSPAR models rely on ad hoc parametrizations for estimating atmospheric density.

\subsubsection{Mars-GRAM}

The Mars Global Reference Atmospheric Model (GRAM) 2010 represents a significant improvement in atmospheric modeling compared to previous models. Mars-GRAM is an engineering-level atmospheric model constructed using input data tables derived from NASA Ames Mars General Circulation Model (MGCM) and the University of Michigan Mars Thermospheric General Circulation Model (MTGCM) outputs. This model has undergone validation against Radio Science and thermal emission spectrometer (TES) data, encompassing both nadir and limb observations. Additionally, it can model potential perturbations in the atmosphere, a feature frequently employed in Monte Carlo simulations for high-fidelity engineering~\cite{justh2014}. 

While Mars-GRAM excels in providing realistic estimates for potential atmospheric density profiles encountered during entry missions, it is usually avoided for online estimation due to its substantial computational overhead. Instead, online estimation approaches commonly use simplified models paired with adaptive techniques, validated through Monte Carlo simulations using truth trajectories propagated using atmospheric density values obtained from Mars-GRAM~\cite{tracy2023}.

\section{Approaches to Density Adaptation} \label{sec:densityadap}

As models cannot provide a perfect estimation of the atmospheric density, various adaptive techniques can be used to address potential errors or unforeseen behavior that the vehicle may encounter during the entry phase. The first approach involves covariance matching, wherein adjustments are made to the process and measurement noise covariance matrices to align with the state and measurement innovations in the navigation filter. The second technique entails the estimation of correction parameters, used to rectify a nominal density profile established during the offline phases of the navigation filter design. Lastly, multiple model adaptive estimation can also be used for this purpose, where a bank of estimators operates in parallel, yielding a more robust estimate by considering different density profiles. This section delves into the presentation and discussion of each of these approaches.

\subsection{Covariance Matching}

Covariance matching techniques have found utility in atmospheric reconstruction due to the inherent uncertainty in atmospheric parameters, where the process and measurement noise covariances matrices are often unknown. When directly estimating the density, that is, the density is part of the estimation state, covariance matching techniques have been shown to be advantageous~\cite{dutta2014}. It is worth noting that incorporating the density as part of the state introduces constraints, notably on the dynamics, as the equation for the rate of change is derived from the hydrostatic equation and the perfect gas law~\cite{dutta2013}.

Mehra~\cite{mehra1972} introduced a covariance matching algorithm yielding a non-unique solution for the estimated covariances. Shortly after, Myers and Tapley~\cite{myers1976} presented an algorithm for providing a unique estimate of noise covariance matrices using batches of state and measurement innovations. To estimate the covariance matrices, the use of the following equations was proposed~\cite{myers1976}. Considering a linear state estimation problem, that is,
\begin{align}
    {\bm{x}}_k & = \mathit{\Phi}_{k,k-1}\bm{x}_{k-1} + \bm{q}_{k-1}, \\
    \bm{y}_k & = H_k{\bm{x}}_k + \bm{\eta}_{k},
\end{align}
where, $\bm{x}$ represents the state, $k$ is the time index, $\mathit{\Phi}_{k,k-1}$ is the state transition matrix that maps from time step $k-1$ to time step $k$, $H_k$ is the state observation matrix, $\bm{q}_{k-1}$ and $\bm{\eta}_{k}$ represent process and measurement noise realizations respectively, the estimated process noise covariance matrix is given by~\cite{myers1976}:
\begin{align}
    \hat{Q} = \frac{1}{N-1} \sum_{i = 1}^N\sq{\pr{\bm{\nu}^{(x)}_i - \bar{\bm{\nu}}^{(x)}_i}\pr{\bm{\nu}^{(x)}_i - \bar{\bm{\nu}}^{(x)}_i}^{\mathrm{T}} - \pr{\frac{N-1}{N}\sq{\mathit{\Phi}_{i,i-1}P^{+}_{i-1}\mathit{\Phi}_{i,i-1}^{\mathrm{T}} - P^{+}_{i}}}}, \label{eq:cm}
\end{align}
where $N$ is the batch size, and $\bm{\nu}^{(x)}$ are the state innovations, such that:
\begin{equation}
    \bm{\nu}^{(x)}_i = \hat{\bm{x}}_i^{+} - \mathit{\Phi}_{i,i-1}\hat{\bm{x}}_{i-1}^{+}.
\end{equation}

The ($+$) represents updated (corrected) estimates and $P_i^+$ is the posterior state error covariance. Similarly, the estimated measurement noise covariance is given by~\cite{myers1976}:
\begin{align}
    \hat{R} = \frac{1}{N-1} \sum_{i = 1}^N\sq{\pr{\bm{\nu}^{(y)}_i - \bar{\bm{\nu}}^{(y)}_i}\pr{\bm{\nu}^{(y)}_i - \bar{\bm{\nu}}^{(y)}_i}^{\mathrm{T}} - \pr{\frac{N-1}{N}\sq{H_iP^{-}_iH_i^{\mathrm{T}}}}},
\end{align}
where $\bm{\nu}^{(y)}$ are the measurement innovations:
\begin{equation}
    \bm{\nu}^{(y)}_i = \bm{y}_i - H_i\hat{\bm{x}}_{i}^{-}.
\end{equation}

In this case, the ($-$) represents propagated (predicted) estimates and $P_i^-$ is the propagated state error covariance. Despite the simplicity of covariance matching for adapting to uncertainty in dynamics and measurement models, it is worth noting that the lack of convergence proof~\cite{mehra1972} can raise concerns, especially if wanted to be used for online estimation in missions requiring high precision, such as Martian entry.

\subsection{State Augmentation and Correction}

Another approach to atmospheric density adaptation includes the correction to a nominal profile by augmenting the estimation state~\cite{tracy2023,zegwe2023}. In this solution, a nominal density profile ($\rho_{\mathrm{nominal}}$) is determined offline and carried by the onboard navigation system. The nominal profile is corrected in an online fashion to account for potential errors. To correct the nominal profile, different approaches can be used. For example, the estimated state can be augmented with a correction factor for the density~\cite{tracy2023},
\begin{equation} \label{eq:ac1}
    \bm{x}_a = \sq{\bm{x} \quad \mathcal{K}}^{\mathrm{T}},
\end{equation}
where:
\begin{equation} \label{eq:ac2}
    \mathcal{K} = \frac{\rho}{\rho_{\mathrm{nominal}}}.
\end{equation}

With this approach, the correction factor can be estimated within the navigation filter and the estimated density can be calculated as:
\begin{equation}
    \hat{\rho} = \hat{\mathcal{K}}\rho_{\mathrm{nominal}}.
\end{equation}

Estimating $\mathcal{K}$ behaves favorably from a numerical perspective, because the estimator primarily seeks to refine an existing approximate density profile rather than directly estimating the density~\cite{tracy2023}. The estimated value tends to hover near one, regardless of altitude, when the nominal density profile closely approximates the actual density experienced by the vehicle. Previous guidance approaches have estimated this correction factor by using the sensed deceleration measurements of the vehicle and applying a low-pass ad hoc filter~\cite{roelke2023}. When a Kalman filter is used for navigation, augmenting the filtering state with this correction factor effectively takes care of applying a low-pass filter.

A different way of correcting a nominal profile involves directly modeling the atmosphere as in \eqref{eq:exponential} and estimating the density at the reference radial position ($\rho_0$)~\cite{levesque2007,zegwe2023} and the atmospheric scale height ($h_s$), thus,
\begin{equation}
    \bm{x}_a = \sq{\bm{x} \quad \rho_0 \quad h_s}.
\end{equation}

However, constraining the nominal density model to an exponential form can be ill-conditioned because the exponential model tends to amplify errors at higher altitudes~\cite{dutta2014}. It is important to note that augmentation approaches perform the adaptation to the atmospheric density as a linear function of the measurement, since the EKF and the UKF are typically used in these scenarios.

\vspace{-1em}\subsection{Others}

Lastly, multiple model adaptive estimation~\cite{zanetti2007,marschke2008} and the use of neural ensembles~\cite{wagner2011} have been proposed for the adaptation of the atmospheric density in Martian entry; these solutions are highly dependent on the number of filters used (in the case of multiple model filtering) or in the number of density models stored in the neural ensemble.

\section{Density Adaptation with Neural Networks} \label{sec:densityadaptwnn}

This work introduces a new approach that uses neural networks within a maximum likelihood framework to adapt to fluctuations in atmospheric density. Previous approaches that have employed maximum likelihood adaptive filtering for tasks like flight-path reconstruction~\cite{chu1996}, typically rely on analytical expressions and Newton-Raphson methods to solve the optimization problem. In contrast, we propose the use of the maximum likelihood approach within a machine learning framework, enabling the use of well-established stochastic optimization methods that have undergone extensive research and refinement~\cite{kingma2017}. This new adaptive filtering technique follows two distinct stages: offline training of a neural network, and its online adaptation for accurate atmospheric density estimation.

\subsection{Offline Training}

First, a neural network is trained to predict atmospheric density as a function of the planet-centric radius, as represented by the equation,
\begin{equation}
\hat{\rho}_{\mathcal{N}\!\mathcal{N}} = \mathcal{N}\!\mathcal{N}(r, \bm{\xi}),
\end{equation}
where the weights and biases of the trained neural network are denoted by $\bm{\xi}$. To train the neural network, synthetic data is generated by simulating Martian entry dynamics using a low-fidelity model of atmospheric density. To create the low-fidelity model, a least-squares exponential fit is obtained from multiple realizations of Mars-GRAM profiles. This fitting process aims to match the data to an exponential model represented as \eqref{eq:exponential}. Using this exponential model, it becomes straightforward to generate synthetic data by simulating various entry trajectories and recording planet-centric radius and atmospheric density at a predefined sampling rate.

Algorithm~\ref{algo:off} provides an overview of the data generation and offline training of the neural network. For this algorithm $N$ trajectories are simulated using the exponential density model and starting from different initial conditions. The planet-centric radius and atmospheric density values are saved at a fixed rate for each trajectory to generate the input and output training data. Section~\ref{subsec:implementation} provides an in-depth description of the offline training implementation.

\begin{algorithm}[htbp]
	\caption{Offline data generation and training} 
    \label{algo:off}
	\begin{algorithmic}[1]
    \State Calculate exponential fit \eqref{eq:exponential} $\hat{\rho}_{\exp}$ from Mars-GRAM profiles
    \State Initialize an input vector as empty $\bm{\mathcal{R}} = []$
    \State Initialize an output vector as empty $\bm{\mathcal{P}}_{\exp} = []$
        \For{$i = 1:N$}
            \State Set initial conditions: $\bm{x}_0$
            \State Propagate initial conditions until a final time $t_f$ using $\hat{\rho}_{\exp}$
            \State Save $r_k$ and $\hat{\rho}_{\exp,k}$ from the trajectory at a fixed rate
            \State Append the sequence of $r_k$ to the input vector $\bm{\mathcal{R}}$
            \State Append the sequence of $\hat{\rho}_{\exp,k}$ to the output vector $\bm{\mathcal{P}}_{\exp}$
        \EndFor
        \State Preprocess input vector $\bm{i} \leftarrow \bm{\mathcal{R}}$
        \State Preprocess output vector $\bm{o} \leftarrow \bm{\mathcal{P}}_{\exp}$
        \State Train neural network from the preprocessed input and output $\pr{\bm{i},\bm{o}}$: $\hat{\rho}_{\mathcal{N}\!\mathcal{N}} = \mathcal{N}\!\mathcal{N}(r, \bm{\xi})$
	\end{algorithmic} 
\end{algorithm}

\subsubsection{Practical implementation} \label{subsec:implementation}

\begin{figure}[h]
    \centering
    \includegraphics[width = 0.5\textwidth]{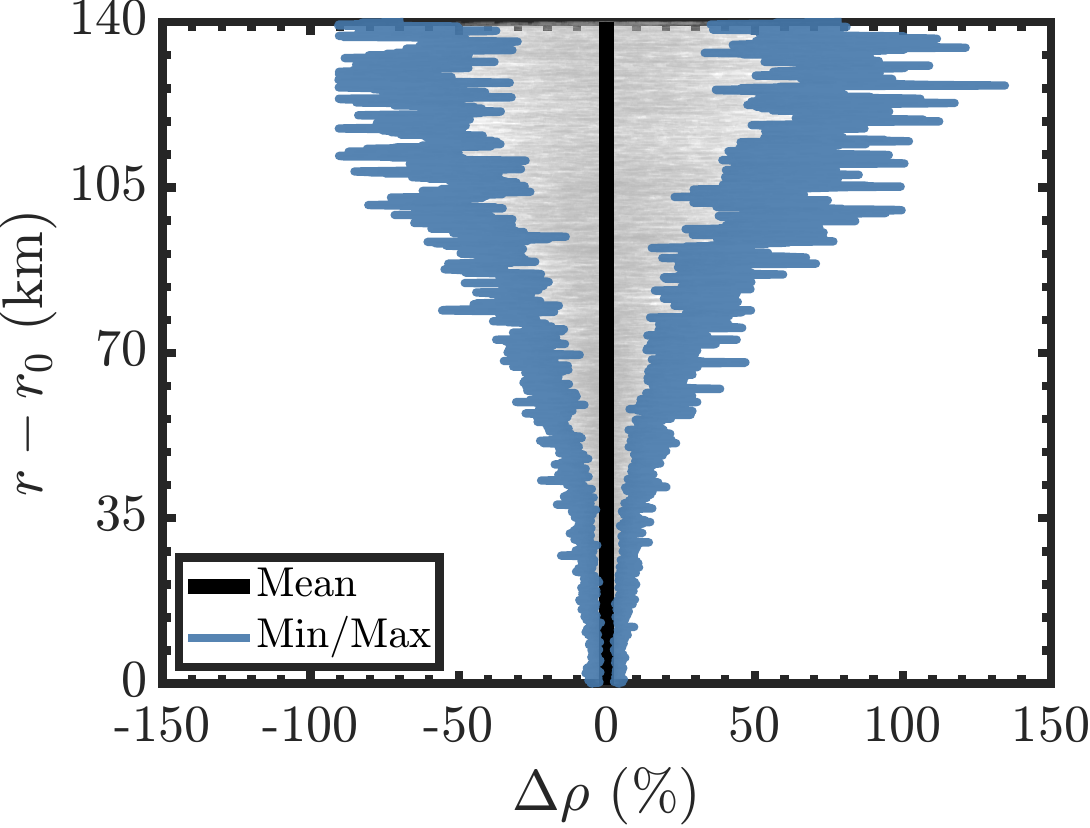}
    \caption{Altitude as a function of the density variation for the Mars-GRAM profiles generated. The gray lines show the different atmospheric profiles.}
    \label{fig:rhogram}
\end{figure}

This section describes the implementation of the offline training of the neural network. It is important to highlight that various implementations can be employed for the neural network, as this network is only a starting point for the online estimation portion. To generate the training data, as previously described, a least squares exponential fit was applied to multiple Mars-GRAM profiles ($\rho_{\mathrm{GRAM}}$). Figure~\ref{fig:rhogram} shows the Mars-GRAM profiles used for the implementation of the neural network. To generate these profiles, the density and wind random perturbations were set to 1, representing a standard deviation in the range of 2\% to 45\% of the unperturbed mean.

\begin{figure}[h]
    \centering
    \includegraphics[width = 0.5\textwidth]{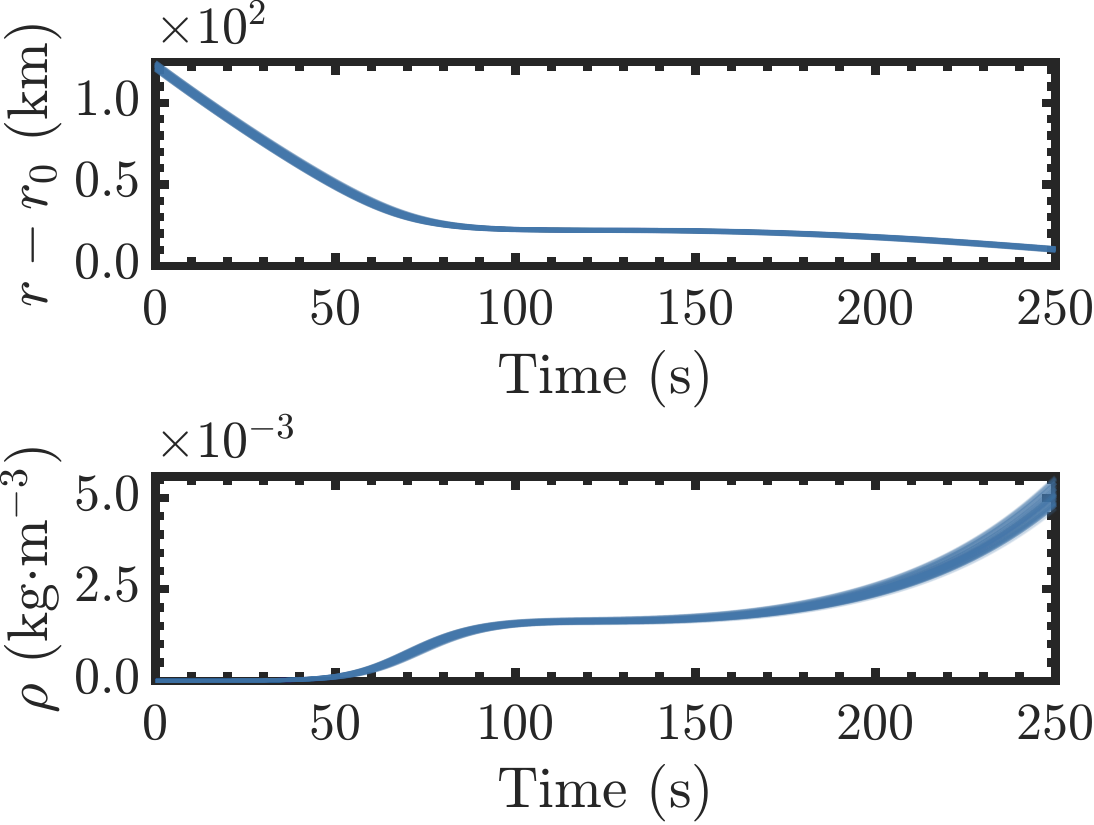}
    \caption{Altitude (top) and density as obtained from the least squares exponential fit (bottom) as a function of time.}
    \label{fig:inputsoutputs}
\end{figure}

Using this exponential fit, 1000 different entry trajectories were simulated based on the dynamic equations outlined in Section~\ref{sec:dynamics}. Each trajectory was initialized with distinct initial conditions sampled from the entry conditions described in Table~\ref{tab:X0-P0}, and data on atmospheric density and planet-centric radius were recorded at 0.5-second intervals during the simulations. The trajectories were propagated for 250 seconds reaching an altitude of approximately 11 kilometers above the Martian surface, a typical point marking the end of the entry phase~\cite{zegwe2023}. Figure~\ref{fig:inputsoutputs} shows different propagated trajectories that use different initial conditions coupled with the exponential fit for the density calculation. 

To train the neural network, a method similar to that introduced by Amato and McMahon~\cite{amato2021} was adopted. To facilitate the training process, the output data, which represents atmospheric density, was transformed using the equation,
\begin{equation}
\varrho_k = \sqrt{\mathcal{B}-\log_{10}\rho_k},
\end{equation}
where $\mathcal{B}$ is a shifting factor ensuring that the argument of the square root is always positive. For this work, $\mathcal{B} = 0$ since the atmospheric density is consistently expressed in SI base units, ensuring that the argument of the square root is always non-negative. To assess the possibility of overfitting, a train/validation split on the data was conducted. Both the training input and output data were normalized, following the transformations,
\begin{align}
i_k & = \frac{r_k - \bar{r}}{\sigma_r}, \\
o_k & = \frac{\varrho_k - \bar{\varrho}}{\sigma_\varrho}.
\end{align}
where $i_k$ is the input to the network and $o_k$ represents the output. Here, $\bar{r}$ represents the mean of all training input data, $\sigma_r$ is the associated standard deviation, $\bar{\varrho}$ is the mean of all training output data, and $\sigma_\varrho$ is the associated standard deviation. With the normalized training data, a simple feed-forward network with one hidden layer containing 100 neurons, and using $\tanh$ as the activation function, was trained. The architecture of this neural network was chosen to accelerate the adaptation process, aiming for a smaller number of hidden layers and neurons, as the weights and biases of the network will be adapted online and recursively.

\begin{figure}[H]
	\centering
	\includegraphics[width = 0.5\textwidth]{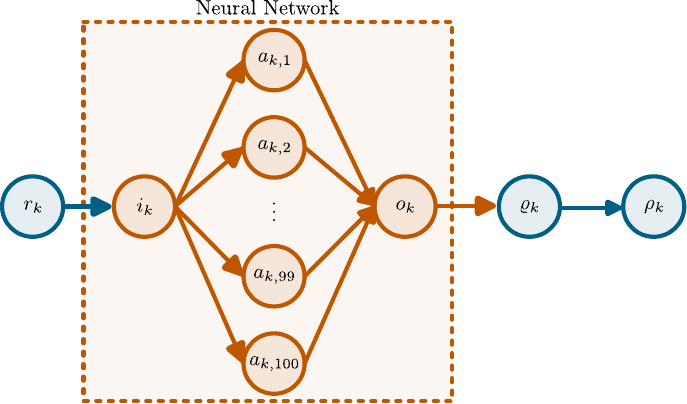}
	\caption{Network architecture used to estimate the atmospheric density from the planet-centric radius.}
	\label{fig:model}
\end{figure}

The network was trained for 1000 epochs using Adam as the optimizer with a varying learning rate ranging from $l_{r_{\min}} = 10^{-6}$ to $l_{r_{\max}} = 10^{-2}$. Figure~\ref{fig:model} illustrates the procedure for estimating atmospheric density using the trained neural network. The planet-centric radius ($r_k$) is first normalized, and this normalized input ($i_k$) is fed into the neural network. Subsequently, the output ($o_k$) is unnormalized to derive the transformed density ($\varrho_k$), and the density ($\rho_k$) is computed using the inverse mapping. The network operations are given by:
\begin{align}
    \bm{a}_k & = \tanh\br{W^{(i)}{i}_k + \bm{b}^{(i)}_k },\\
    \bm{o}_k & = W^{(o)}\bm{a}_k + \bm{b}^{(o)}_k,
\end{align}
where $\bm{a}_k \in \mathbb{R}^{100}$ are the neurons in the hidden layer, $W^{(i)} \in \mathbb{R}^{100 \times 1}$ is the vector of weights for the first layer, $i_k \in \mathbb{R}$ represents the input and $\bm{b}^{(i)}_k \in \mathbb{R}^{100}$ are the biases for the first layer. Similarly, $o_k \in \mathbb{R}$ is the output of the network, and $W^{(o)} \in \mathbb{R}^{1 \times 100}$ and $\bm{b}^{(o)}_k \in \mathbb{R}$ are the weights and biases for the output layer, respectively. Note that the time complexity of a forward pass through the network is $\mathcal{O}(1 \cdot h_0 + h_0 \cdot 1) = \mathcal{O}(2\cdot h_0)$ with $h_0$ the number of neurons (in this case 100).

\begin{figure}[H]
    \centering
    \includegraphics[width = 0.5\textwidth]{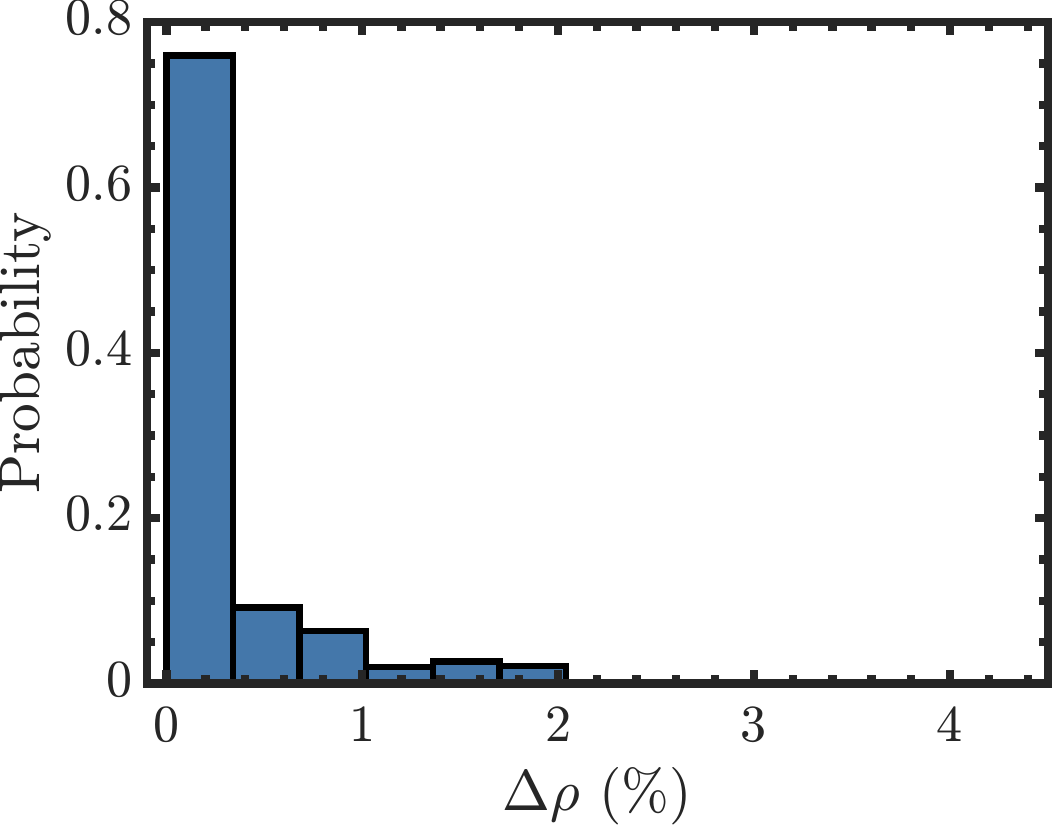}
    \caption{Probability histogram of the relative error between the predicted density from the trained neural network and the density in the validation set.}
    \label{fig:nnpred}
\end{figure}

Figure~\ref{fig:nnpred} displays the percentage error between the estimated density from the trained neural network and the actual density in the validation set. As it can be seen, the network accurately predicts the density, as most errors are less than 1\%, suggesting a good fit to the exponential model. It is important to note that the accuracy of this network is not crucial for the proposed approach. The network is expected to perform poorly on real atmospheric density profiles, which is why the weights and biases will be updated online to adapt to a higher fidelity density model. However, achieving a good fit to an exponential model is desirable.

\subsection{Online Estimation}

Once the neural network has been trained, an online adaptation scheme can be applied. For this work, the state to be estimated is defined as:
\begin{equation}
\bm{x} = \left[r \quad \phi \quad \theta \quad v \quad \phi \quad \psi \quad B \quad L/D\right]^T,
\end{equation}
where the inverse of the ballistic coefficient ($B$) is modeled as a random walk, employing a similar approach for modeling the lift-to-drag ratio ($L/D$)~\cite{zegwe2023}. The available measurement vector is expressed as:
\begin{equation}
\bm{y} = \left[\tilde{\bm{a}}^b \quad \tilde{q} \quad \tilde{\dot{Q}}_s\right].
\end{equation}

To simplify notation, the dynamics and measurement models are defined as:
\begin{align}
\dot{\bm{x}} & = f(\bm{x}, \rho), \label{eq:continuousdyn}\\
\bm{y}_k & = h\pr{\bm{x}_k, \rho_k} + \bm{\eta_k},
\end{align}
where $f$ are the dynamics presented in section~\ref{sec:dynamics}, $h$ refers to the measurement model in section~\ref{sec:measurements}, $\rho$ is the atmospheric density and $\bm{\eta_k}$ represents measurement noise. Process noise is modeled in the solution of the dynamics by,
\begin{equation}
    \bm{x}_k = F\pr{\bm{x}_{k-1},\rho_{k-1}} + \bm{q}_{k-1}
\end{equation}
where $F$ is the flow of \eqref{eq:continuousdyn} and $\bm{q}_{k-1}$ is the process noise. For this scenario, the selected online estimator is the unscented Schmidt-Kalman filter (USKF)~\cite{zanetti2013,stauch2015}. The USKF is a linear update filter known for its robustness in handling nonlinear dynamics and measurements. This choice eliminates the need for partial derivatives both during time propagation and measurement updates, which can otherwise lead to computational slowdown as the neural network is being used to approximate the density, thus requiring the need of more complicated partial derivatives. Additionally, this filter incorporates a consider analysis to account for the uncertainty in the atmospheric density estimates, as this quantity is not part of the filter's state space.

\subsubsection{Time propagation}

Each iteration starts from the previous time posterior estimate of the state $\hat{\bm{x}}^+_{a,k-1}$ and covariance $P^{+}_{a,k-1}$, where both the estimate and covariance are augmented to include the consider parameter,
\begin{equation}
    \hat{\bm{x}}^{+}_{a,k-1} = \begin{bmatrix} \hat{\bm{x}}^{+\mathrm{T}}_{k-1} & \hat{\bm{c}}^{+}_{k-1} \end{bmatrix}^{\mathrm{T}},
\end{equation}
\begin{equation}
    P^{+}_{a,k-1} = \begin{bmatrix}
      P^{+}_{k-1} & C^{+}_{k-1} \\
      \left(C^{+}_{k-1}\right)^\mathrm{T} & P^{+}_{c,k-1}
    \end{bmatrix},
\end{equation}
where the initial value $P^{+}_{c,0}$ is user-defined (note that $\hat{\bm{c}}^{+}_{0} = 1$ and $C^{+}_0 = 0_{n_x \times 1}$). With these augmented state and covariance, a set of posterior estimates is built by calculating the associated sigma points~\cite{julier2004},
\begin{align}\label{eq:sigmapointsbegin}
        \hat{\bm{\mathcal{X}}}_{k-1}^{+(0)} & = \hat{\bm{x}}^{+}_{a,k-1}, \\
       \hat{\bm{\mathcal{X}}}_{k-1}^{+(i)} & = \hat{\bm{x}}^{+}_{a,k-1}  + \left(\sqrt{L+\lambda P^{+}_{a,k-1}}\right)^{(i)}, \hspace{1em} i = 1, \cdots , L, \\
        \hat{\bm{\mathcal{X}}}_{k-1}^{+(i)} & = \hat{\bm{x}}^{+}_{a,k-1}  - \left(\sqrt{L+\lambda P^{+}_{a,k-1}}\right)^{(i)}, \hspace{1em} i = L+1, \cdots , 2L, \label{eq:sigmapointsend}
\end{align}
resulting in a set of $2L + 1$ samples. The superscript $(i)$ denotes the $i$-th column of the matrix in parenthesis. The scaling parameter $\lambda$ is defined as $\lambda = \alpha^2\left(L + \kappa \right) - L$, where $\alpha$ and $\kappa$ are secondary scaling parameters and $L = n_x + 1$. The parameter $\alpha$ controls the spread of points around $\hat{\bm{x}}^{+}_{a,k-1}$ and is typically set to a small positive value ($10^{-4} \leq \alpha \leq 1$), and $\kappa$ is usually chosen as $3-L$~\cite{julier2004}. 

After calculating the sigma points, these can be propagated to the next time step by integrating the dynamics and using the trained neural network to estimate the density. With $F_a(\bm{x},\rho)$ representing both the flow of $f(\bm{x},\rho)$ and the flow of the consider parameter modeled as an exponentially correlated random variable (ECRV)~\cite{zanetti2012} centered at one, the states are propagated by (the operator $\mathcal{N}\!\mathcal{N}\pr{\cdot}$ handles the normalization and transformation steps):
\begin{align}
    \bm{\hat{\mathcal{X}}}^{-(i)}_{k} & = F_a\left({\hat{\bm{\mathcal{X}}}}^{+(i)}_{k-1},\mathcal{P}^{(i)}_{k-1}\right), \\
    \mathcal{P}^{(i)}_{k-1} & = \hat{\bm{\mathcal{X}}}^{(i)+}_{k-1}(n_x+1)\cdot\mathcal{N}\!\mathcal{N}\pr{\hat{\bm{\mathcal{X}}}^{+(i)}_{k-1}(1),\bm{\xi}_{k-1}},
\end{align}
where $\hat{\bm{\mathcal{X}}}^{(i)+}_{k-1}(n_x+1)$ is used as Matlab notation and represents the consider parameter at each sigma point, $\hat{\bm{\mathcal{X}}}^{+(i)}_{k-1}(1)$ is the estimated planet-centric radius at each sigma point and $\bm{\xi}_{k-1}$ are the current weights and biases of the neural network. Note that the consider parameter is dispersing the solution of the neural network to account for the uncertainty in the atmospheric density estimate. With the propagated sigma points, a prior estimate and covariance can be calculated by a weighted sum, such that:
\begin{align}
    \hat{\bm{x}}^{-}_{a,k} & = \sum_{i = 0}^{2L} w_m^{(i)}\hat{\bm{\mathcal{X}}}^{-(i)}_k, \\
    P^-_{a,k} & = \sum_{i = 0}^{2L} w_c^{(i)}\sq{\hat{\bm{\mathcal{X}}}^{-(i)}_k - \hat{\bm{x}}^{-}_k}\sq{\hat{\bm{\mathcal{X}}}^{-(i)}_k - \hat{\bm{x}}^{-}_k}^{\mathrm{T}} + Q_{a,k-1},
\end{align}
where the weights are given by~\cite{wan2000}:
\begin{align}
        w_m^{(0)} & = \frac{\strut\lambda}{\strut L + \lambda},  \label{eq:sigmawbegin}\\
        w_c^{(0)} & = \frac{\strut\lambda}{\strut L + \lambda}  + 1 - \alpha^2 + \beta,  \\
        w_m^{(i)} & = w_c^{(i)} = \frac{\strut 1}{\strut 2\left(L + \lambda\right)}, \hspace{1.5em} i = 1, ..., 2L, \label{eq:sigmawend}
\end{align}
and the augmented process noise covariance is defined as:
\begin{equation}
    Q_{a,k-1} = \begin{bmatrix}
     Q_{k-1} & 0_{n_x \times 1} \\
      0_{1 \times n_x} & Q_{c,k-1}
    \end{bmatrix},
\end{equation}
with $Q_{c,k-1} = \pr{1-\exp\br{-2\Delta t_k/\tau}}P_{ss}$, where $\Delta t_k$ is the propagation interval, $\tau$ is the time constant of the ECRV and $P_{ss}$ is the steady-state covariance of this variable. The constant $\beta$ is used to incorporate prior knowledge of the probability distribution. Under the assumption of a scalar and Gaussian prior distribution, $\beta = 2$ is usually regarded as optimal since it can cancel second-order error terms~\cite{wan2001}. From the augmented state and covariance, the estimation state and its covariance can be retrieved by selecting the first $n_x$ elements of the resulting vector and the $n_x$ by $n_x$ upper block of the resulting matrix. 

\vspace{-1em}\subsubsection{Maximum likelihood optimization}

Once a measurement ($\bm{y}_k$) is obtained, a loss function is constructed using the measurement log likelihood,
\begin{align}
    \mathcal{L}\pr{\xi_{k-1}} & = \sq{\bm{y}_k - h\pr{\hat{\bm{x}}^{-}_k,\hat{\rho}_{k-1}}}^{\mathrm{T}}R^{-1}\sq{\bm{y}_k - h\pr{\hat{\bm{x}}^{-}_k,\hat{\rho}_{k-1}}}, \label{eq:loss}\\
    \hat{\rho}_{k-1} & = \mathcal{N}\!\mathcal{N}\pr{\hat{\bm{x}}^{-}_k(1),\bm{\xi}_{k-1}}.
\end{align}

This loss function highlights the fact that the adaptation is performed in a nonlinear fashion with respect to the measurement, as opposed to the previous augmentation approaches described in section~\ref{sec:densityadap}. Thus, a minimization problem can be constructed, where,
\begin{equation} \label{eq:minproblem}
    \bm{\xi}_k = \arg\min_{\xi_{k-1}}\mathcal{L}\pr{\xi_{k-1}}.
\end{equation}

When dealing with non-convex problems, fine-tuning the learning rate in traditional optimization algorithms (such as Newton-Raphson) can present challenges and may lead to divergence issues. Instead, we propose the use of automatic differentiation and stochastic optimizers to address the minimization problem in \eqref{eq:minproblem}. Given that this problem pertains to the optimization of the weights and biases of a neural network, back-propagation (also known as adjoint-mode differentiation) can be employed to compute gradients up to floating-point precision. Furthermore, the adaptive and stochastic optimization capabilities offered by the Adam optimizer~\cite{kingma2017} can be leveraged to minimize the loss function. A notable advantage of using the Adam optimizer is its ability to incorporate memory of previous gradients from earlier time steps, thereby assisting in the fine-tuning of the adaptive learning rate within the optimizer. Starting from an initial gradient $\nabla_{\bm{\xi}_{k-1}}\mathcal{L}$, an initial moving average of the gradient $\bar{\nabla}_{k-1}$, and an initial moving average of the squared gradient $\pr{\bar{\nabla}_{k-1}}^2$, where the notation $\pr{\cdot}^2$ indicates element-wise square, Adam iteratively updates the network parameters as follows. First the moving average of the gradient and the gradient squared are updated,
\begin{align}
    \bar{\nabla}_{k-1}^{\star} & = \beta_1\bar{\nabla}_{k-1} + \pr{1-\beta_1}\nabla_{\bm{\xi}_{k-1}}\mathcal{L}, \\
    \pr{\bar{\nabla}_{k-1}^{\star}}^2 & = \beta_2\pr{\bar{\nabla}_{k-1}}^2 + \pr{1-\beta_2}\pr{\nabla_{\bm{\xi}_{k-1}}\mathcal{L}}^2,
\end{align}
where $\beta_1$ and $\beta_2$ are exponential decay rates of the moving averages being estimated. Once these two estimates are obtained, the network parameters are updated using the following equation:
\begin{equation}
    \bm{\xi}_{k-1}^\star = \bm{\xi}_{k-1} - \alpha_{\mathcal{L}}\pr{{\bar{\nabla}_{k-1}^{\star}}\Big/\sq{{\sqrt{\pr{\bar{\nabla}_{k-1}^{\star}}^2}+\varepsilon}}}.
\end{equation}

Here, $\alpha_{\mathcal{L}}$ represents the step size, and $\varepsilon$ is a small constant used to prevent numerical overflow. This process repeats until a convergence criterion is met. When a convergence criterion is satisfied, the updated network parameters and moving averages are saved for the measurement update and the next iteration. Thus, upon reaching convergence,
\begin{align}
    \bar{\nabla}_k & = \bar{\nabla}_{k-1}^{\star}, \\
    \pr{\bar{\nabla}_k}^2 & = \pr{\bar{\nabla}_{k-1}^{\star}}^2,\\
     \bm{\xi}_{k} & =  \bm{\xi}_{k-1}^\star.
\end{align}

For this work, the moving average of the gradient was initialized to zero, and the square of the gradient was initialized to the current gradient squared. Additionally, the moving averages were not corrected for the initialization bias. These design strategies were informed by the performance of the algorithm when deployed online in different filtering tests. In practice, the maximization step is only performed if the evaluated loss exceeds a predetermined threshold ($\mathcal{T}$). This helps alleviate the computational burden of the online filter and leads to less noisy outcomes. Algorithm~\ref{algo:mlo} provides a comprehensive overview of the maximum likelihood optimization step. In this algorithm, a maximum patience value $(p_{\max})$ is defined to prevent the degradation of the neural network in the event of encountering unfavorable gradients.

\begin{algorithm}[h]
	\caption{Maximum likelihood optimization (MLO)} 
    \label{algo:mlo}
	\begin{algorithmic}[1]
        \Statex{\hspace{-1.5em} \textbf{Input:} $\hat{\bm{x}}^{-}_{a,k}, P^{-}_{a,k},\bm{\xi}_{k-1},\bar{\nabla}_{k-1},\pr{\bar{\nabla}_{k-1}}^2,\alpha_{\mathcal{L}},\beta_1,\beta_2,p_{\max},\mathcal{T}$}
        \Statex{\hspace{-1.5em} \textbf{Output:} $\bm{\xi}_{k},\bar{\nabla}_{k},\pr{\bar{\nabla}_{k}}^2$}
        \State Calculate initial loss \eqref{eq:loss}: $\mathcal{L}(\bm{\xi}_{k-1})$
        \State Initialize patience: $p = 0$
        \While {$\mathcal{L}(\bm{\xi}_{k-1})$ > $\mathcal{T}$}
            \State Apply back-propagation to obtain the gradients: $\nabla_{\bm{\xi}_{k-1}}\mathcal{L}$
            \State Use Adam to obtain new parameters, average gradient and average gradient squared:
            \State $\sq{\bm{\xi}^{\star}_{k-1},\bar{\nabla}^{\star}_{k-1},\pr{\bar{\nabla}^{\star}_{k-1}}^2}$ = Adam$\sq{\bm{\xi}_{k-1},\nabla_{\bm{\xi}_{k-1}}\mathcal{L},\bar{\nabla}_{k-1},\pr{\bar{\nabla}_{k-1}}^2,\alpha_{\mathcal{L}},\beta_1,\beta_2}$
            \State Evaluate new loss \eqref{eq:loss}: $\mathcal{L}(\bm{\xi}^{\star}_{k-1})$
            \If{$\mathcal{L}(\bm{\xi}^{\star}_{k-1})$ < $\mathcal{L}(\bm{\xi}_{k-1})$}
            \State Update adapted parameters: $\bm{\xi}_{k-1} \leftarrow \bm{\xi}^{\star}_{k-1}$
            \State Update Adam parameters: $\pr{\bar{\nabla}_{k-1},\pr{\bar{\nabla}_{k-1}}^2} \leftarrow \pr{\bar{\nabla}_{k-1}^{\star},\pr{\bar{\nabla}^{\star}_{k-1}}^2}$
            \State Update loss: $\mathcal{L}(\bm{\xi}_{k-1}) \leftarrow \mathcal{L}(\bm{\xi}^{\star}_{k-1})$
            \Else
            \State Update patience: $p = p + 1$
            \EndIf
            \If{$p = p_{\max}$}
            \State \textbf{break}
            \EndIf
        \EndWhile
        \State Save adapted parameters: $\bm{\xi}_{k} \leftarrow \bm{\xi}_{k-1}$
        \State Save Adam parameters: $\pr{\bar{\nabla}_{k},\pr{\bar{\nabla}_{k}}^2}  \leftarrow \pr{\bar{\nabla}_{k-1},\pr{\bar{\nabla}_{k-1}}^2}$
	\end{algorithmic} 
\end{algorithm}

As previously mentioned, the time complexity of a forward pass through the network is $\mathcal{O}(2\cdot h_0)$. Assuming that the back-propagation shares the same order of magnitude in time complexity as the forward pass, the total time complexity for one optimization step is $\mathcal{O}(2\cdot h_0)$. The time complexity scales linearly with the number of optimization steps $(n_o)$, resulting in $\mathcal{O}(2\cdot h_0 \cdot n_o)$. Therefore, careful consideration of the number of optimization steps is important to ensure affordable online implementation. Nonetheless, these time complexity considerations lie outside the scope of the current research, as the primary goal of this work is to demonstrate the capabilities of using neural networks in an adaptive manner. 

\subsubsection{Measurement update}

Once the adapted weights and biases of the network have been obtained, the measurement update can be performed. To do this, sigma points are calculated around the prior estimate by using the propagated state and covariance and using the same approach as in equations \eqref{eq:sigmapointsbegin}-\eqref{eq:sigmapointsend}. With the set of prior sigma points, a set of expected measurements is calculated by using the measurement model and the adapted weights and biases of the network,
\begin{align}
    \hat{\mathcal{Y}}^{(i)}_k & = h\pr{\hat{\bm{\mathcal{X}}}^{-(i)}_k,\mathcal{P}^{(i)}_k}, \\
    \mathcal{P}^{(i)}_k & = \hat{\bm{\mathcal{X}}}^{-(i)}_k(n_x + 1)\cdot\mathcal{N}\!\mathcal{N}\pr{\hat{\bm{\mathcal{X}}}^{-(i)}_k(1),\bm{\xi}_k}.
\end{align}

Notice again that the atmospheric density estimate obtained from the network is slightly perturbed by the consider parameter for each sigma point in order to account for uncertainty. With the set of expected measurements, the predicted mean, its covariance and cross-covariance with the state can be calculated to perform a Kalman update. Using a similar weighted sum as in the time propagation,
\begin{align}
    \hat{\bm{y}}_k & = \sum_{i = 0}^{2L} w_m^{(i)}\hat{\mathcal{Y}}^{(i)}_k, \\
    P_{yy,k} & = \sum_{i = 0}^{2L} w_c^{(i)}\sq{\hat{\bm{\mathcal{Y}}}^{(i)}_k - \hat{\bm{y}}_k}\sq{\hat{\bm{\mathcal{Y}}}^{(i)}_k - \hat{\bm{y}}_k}^{\mathrm{T}} + R_k, \\
    P_{x_ay,k} & = \sum_{i = 0}^{2L} w_c^{(i)}\sq{\hat{\bm{\mathcal{X}}}^{-(i)}_k - \hat{\bm{x}}^{-}_k}\sq{\hat{\bm{\mathcal{Y}}}^{(i)}_k - \hat{\bm{y}}_k}^{\mathrm{T}},
\end{align}
where $R_k$ is the measurement noise covariance, and the weights are calculated using the same equations as in the time propagation. It is important to note that for the measurement update, the constants $\alpha,\beta$ and $\kappa$ do not necessarily have to be the same as in the time propagation step. With the obtained matrices, a Kalman gain is calculated,
\begin{equation} \label{eq:Kk}
    K_k = P_{x_ay,k}P_{yy,k}^{-1} = \begin{bmatrix}K_{x,k} \\ K_{c,k}\end{bmatrix}.
\end{equation}

Finally, a Kalman update can be performed to obtain the posterior estimate and covariance, such that:
\begin{align}
    \hat{\bm{x}}^+_{a,k} & = \hat{\bm{x}}^-_{a,k} + \begin{bmatrix}K_{x,k} \\ 0\end{bmatrix}\pr{\bm{y_k} - \hat{\bm{y}}_k}, \label{eq:posteriorstate} \\
    P^{+}_{a,k} & = P^{-}_{a,k} - \begin{bmatrix} K_{x,k}P_{yy,k}K_{x,k}^{\mathrm{T}} & K_{x,k}P_{yy,k}K_{c,k}^{\mathrm{T}} \\ K_{c,k}P_{yy,k}K_{x,k}^{\mathrm{T}} & 0 \end{bmatrix}. \label{eq:posteriorcov}
\end{align}

Just as with the time propagation, from the augmented state and covariance, the estimation state and its covariance can be retrieved by selecting the first $n_x$ elements of the resulting vector and the $n_x$ by $n_x$ upper block of the resulting matrix. It is important to note that the consider parameter and its associated covariance are not updated in this filter.

\begin{figure}[h]
    \centering   
    \includegraphics[width = 0.5\textwidth]{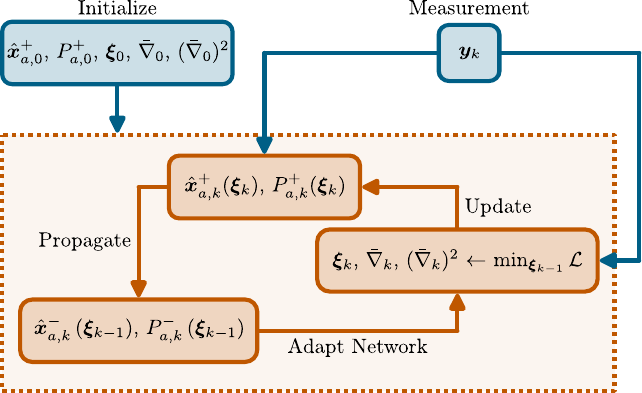}
    \caption{Flowchart of the online estimation scheme.}
    \label{fig:flow}
\end{figure}

Algorithm~\ref{algo:onlineest} shows the full recursion algorithm for the online estimation. Initially, the neural network is trained offline to obtain the initial weights and biases ($\bm{\xi}_0$) and the initial Adam parameters ($\bar{\nabla}_0$, $(\bar{\nabla}_0)^2$). In addition, the initial state and covariance ($\hat{\bm{x}}^{+}_{a,0}$, $P^{+}_{a,0}$) are set as the starting point ($k-1$). Then, the state and covariance are propagated ($\hat{\bm{x}}^{-}_{a,k}$, $P^{-}_{a,k}$) using the current network. Once the measurement is obtained $(\bm{y}_k)$, the loss function $(\mathcal{L})$ is minimized using Adam and the network ($\bm{\xi}_k$) and optimizer parameters ($\bar{\nabla}_k$, $(\bar{\nabla}_{k})^2$) are adapted. Finally, the state and covariance are updated ($\hat{\bm{x}}^{+}_{a,k}$, $P^{+}_{a,k}$) using the adapted network. A summary of the complete online estimation scheme is also presented in Fig.~\ref{fig:flow} as a flowchart.

\begin{algorithm}[h]
	\caption{Recursion algorithm} 
    \label{algo:onlineest}
	\begin{algorithmic}[1]
        \State Initialize estimation parameters: $\hat{\bm{x}}^{+}_{a,0}, P^{+}_{a,0}$
        \State Initialize neural network parameters: $\bm{\xi}_{0},\bar{\nabla}_{0},\pr{\bar{\nabla}_{0}}^2$
        \State Declare optimization variables: $\alpha_{\mathcal{L}},\beta_1,\beta_2,p_{\max},\mathcal{T}$
        \State Declare unscented transform constants: $\alpha,\beta,\kappa$
        \While {$\bm{y}_k$ is available}
        \LComment{Time propagation}
        \State Calculate sigma points \eqref{eq:sigmapointsbegin}-\eqref{eq:sigmapointsend} $\hat{\bm{\mathcal{X}}}^{+}_{k-1}$ from $\hat{\bm{x}}^{+}_{a,k-1}, P^{+}_{a,k-1}$
        \State Propagate sigma points to current time step: $\bm{\hat{\mathcal{X}}}^{-}_{k} = F_a\left({\hat{\bm{\mathcal{X}}}}^{+}_{k-1},\mathcal{P}_{k-1}\right)$
        \State Calculate sigma points weights \eqref{eq:sigmawbegin}-\eqref{eq:sigmawend}: $w_m^{(i)},w_c^{(i)}$
        \State Reduce sigma points:
        \State $\hat{\bm{x}}^{-}_{a,k} = \sum_{i = 0}^{2L} w_m^{(i)}\hat{\bm{\mathcal{X}}}^{-(i)}_k$
        \State $P^-_{a,k} = \sum_{i = 0}^{2L} w_c^{(i)}\sq{\hat{\bm{\mathcal{X}}}^{-(i)}_k - \hat{\bm{x}}^{-}_k}\sq{\hat{\bm{\mathcal{X}}}^{-(i)}_k - \hat{\bm{x}}^{-}_k}^{\mathrm{T}} + Q_{a,k-1}$
        \Statex
        \LComment{Maximum likelihood optimization (Algorithm~\ref{algo:mlo})}        \State$\left[\bm{\xi}_{k},\bar{\nabla}_{k},\pr{\bar{\nabla}_{k}}^2\right] = \mathrm{MLO}\sq{\hat{\bm{x}}^{-}_{a,k}, P^{-}_{a,k},\bm{\xi}_{k-1},\bar{\nabla}_{k-1},\pr{\bar{\nabla}_{k-1}}^2,\alpha_{\mathcal{L}},\beta_1,\beta_2,p_{\max},\mathcal{T}}$
        \Statex
        \LComment{Measurement update}
        \State Calculate sigma points \eqref{eq:sigmapointsbegin}-\eqref{eq:sigmapointsend} $\hat{\bm{\mathcal{X}}}^{-}_{k}$ from $\hat{\bm{x}}^{-}_{a,k}, P^{-}_{a,k}$
        \State Calculate expected measurements: $\hat{\mathcal{Y}}_k = h\pr{\hat{\bm{\mathcal{X}}}^{-}_k,\mathcal{P}_k}$
        \State Calculate sigma points weights \eqref{eq:sigmawbegin}-\eqref{eq:sigmawend}: $w_m^{(i)},w_c^{(i)}$
        \State Reduce sigma points:
        \State $\hat{\bm{y}}_k = \sum_{i = 0}^{2L} w_m^{(i)}\hat{\mathcal{Y}}^{(i)}_k$
        \State $P_{yy,k} = \sum_{i = 0}^{2L} w_c^{(i)}\sq{\hat{\bm{\mathcal{Y}}}^{(i)}_k - \hat{\bm{y}}_k}\sq{\hat{\bm{\mathcal{Y}}}^{(i)}_k - \hat{\bm{y}}_k}^{\mathrm{T}} + R_k$
        \State $P_{x_ay,k} = \sum_{i = 0}^{2L} w_c^{(i)}\sq{\hat{\bm{\mathcal{X}}}^{-(i)}_k - \hat{\bm{x}}^{-}_k}\sq{\hat{\bm{\mathcal{Y}}}^{(i)}_k - \hat{\bm{y}}_k}^{\mathrm{T}}$
        \State Calculate Kalman gain \eqref{eq:Kk}: $K_k$ 
        \State Calculate posterior estimate and covariance \eqref{eq:posteriorstate}-\eqref{eq:posteriorcov}: $\hat{\bm{x}}^{+}_{a,k}, P^{+}_{a,k}$
        \State Save estimation parameters for next measurement: $\pr{\hat{\bm{x}}^{+}_{a,k-1}, P^{+}_{a,k-1}} \leftarrow \pr{\hat{\bm{x}}^{+}_{a,k}, P^{+}_{a,k}}$
        \State Save neural network parameters for next measurement: $\pr{\bm{\xi}_{k-1},\bar{\nabla}_{k-1},\pr{\bar{\nabla}_{k-1}}^2} \leftarrow \pr{\bm{\xi}_{k},\bar{\nabla}_{k},\pr{\bar{\nabla}_{k}}^2} $
        \EndWhile
	\end{algorithmic} 
\end{algorithm}

\section{Results and Discussion} \label{sec:resultsanddiscussion}

To assess the density adaptation scheme, a comprehensive Monte Carlo testing approach was employed. This involved conducting 1000 Monte Carlo runs to thoroughly evaluate the robustness and performance of the algorithms presented. Each true trajectory was simulated using an independent and distinct Mars-GRAM testing trajectory, different from the set previously mentioned, which was used for obtaining the least squares exponential fit and training the network. For each new Mars-GRAM density profile, the random perturbations were also set to 1. Given that Mars-GRAM trajectories provide density at discrete altitudes, trajectory points were interpolated using a cubic spline to determine density values at the query altitude.

\subsection{Monte Carlo Settings} 

All trajectories were initialized with the entry conditions for the Mars Science Laboratory (MSL), outlined in Table~\ref{tab:X0-P0}. This implies simulating potential MSL entry trajectories for various densities the vehicle might encounter. It is crucial to note that these trajectories are simulated in an open-loop fashion, meaning that no bank control is executed, as the focus of this work is solely on the navigation aspect. A constant bank angle ($\sigma = 0$) and a constant angle of attack ($\alpha = -17^\circ$) were assumed for all trajectories, based on the available initial entry conditions of the MSL. 

\begin{table}[H]
    \centering
    \caption{Entry conditions for the true trajectories}
    \begin{tabular}{lcc}
    \hline \hline
       Parameter  & Value & Standard Deviation (3$\sigma$) \\ \hline  
       $r_0$ (m)~\cite{dutta2014b} & \num{3.5222 e6} & \num{3.2066 e1}\\ 
       $\phi_0$ (deg)~\cite{dutta2014b} & \num{-0.3919e1} & \num{7.8100e-4} \\
       $\theta_0$ (deg)~\cite{dutta2014b} & \num{1.2672e2} & \num{3.6700e-4} \\
        $v_0$ (m/s)~\cite{dutta2014b} & \num{6.0833e3} & \num{2.6059e-2} \\
       $\gamma_0$ (deg)~\cite{dutta2014b} & \num{-1.5489e1} & \num{4.0000e-4}\\
       $\psi_0$ (deg)~\cite{dutta2014b} & \num{9.3206e1} & \num{2.6800e-4}\\
       $B_0$ (m$^2$/kg) & \num{7.1000e-3}~\cite{way2006} & \num{4.8000e-3}~\cite{zegwe2023}\\
       $L_0/D_0$ (n.d.) & \num{2.4000e-1}~\cite{way2006} & \num{1.5178e-1}~\cite{zegwe2023}\\\hline \hline
    \end{tabular}
    \label{tab:X0-P0}
\end{table}

The true dynamics were propagated for 350 seconds, reaching altitudes close to 11 kilometers. In addition, the dynamics were propagated with additive zero-mean Gaussian process noise and measurements were calculated using the true state with additive zero-mean Gaussian noise as well. Both the process noise and measurement noise were assumed to be white and uncorrelated with any additional sources of error. For this work, all sensors were sampled at a uniform frequency of 4 Hz. Tables~\ref{tab:Q} and~\ref{tab:R} provide the statistics used for the process and measurement noise, respectively.

\begin{table}[H]
    \centering
    \caption{Process noise statistics}
    \begin{tabular}{lc}
    \hline \hline
       Parameter  & Standard Deviation (\num{3}$\sigma$) \\ \hline  
       $r$ (m)& - \\ 
       $\phi$ (deg) & - \\
       $\theta$ (deg) & - \\
       $v$ (m/s) & \num{3e-1}\\
       $\gamma$ (deg) & \num{2e-3}\\
       $\psi$ (deg) & \num{2e-4}\\
       $B$ (m$^2$/kg) & \num{1e-5}\\
       $L/D$ (n.d.) & \num{3e-5} \\\hline \hline
    \end{tabular}
    \label{tab:Q}    
\end{table}

\begin{table}[H]
    \centering
    \caption{Measurement noise statistics}
    \begin{tabular}{lc}
    \hline \hline
       Parameter  & Standard Deviation (\num{3}$\sigma$) \\ \hline 
       $\tilde{\bm{a}}^b$ $(\mu g)$~\cite{brown2012} & \num{3e2} \\
       $\tilde{q}$ (\% of reading)~\cite{gazarik2008} & \num{1}\\
       $\tilde{\dot{Q}}_s$  (\% of reading) & \num{1}\\ \hline \hline
    \end{tabular}
    \label{tab:R}
\end{table}

\subsection{Filtering Algorithms} 

To evaluate the effectiveness of the density adaptation scheme, three filters, each with different adaptive techniques, were implemented for comparative analysis. It is important to note that the focus of the comparison lies on the available adaptive methods that can be utilized online to account for the mismatch between the onboard atmospheric model and the true atmospheric density. Therefore, for a fair comparison, all adaptive methods start from the same base model, in this case the least squares exponential fit to the MARS-GRAM training trajectories. Although the exponential model is known to be a poor approximation to the true density, as it can amplify errors at higher altitudes~\cite{dutta2014}, the adaptive methods are expected to account for the approximation mismatches. In addition, each solution uses identical base state spaces and measurements to enable meaningful comparison. This approach ensures a fair assessment of the strategies under consideration.

\begin{itemize}
    \item The first filter, labeled UKF-CM (CM for covariance matching), uses the least squares exponential fit to the Mars-GRAM training trajectories for density estimation. A covariance matching technique is used to estimate the process noise covariance as defined in \eqref{eq:cm}, with a batch size of 10 time steps, following the approach presented in~\cite{dutta2014}. Since the estimate of the process noise covariance may become negative definite, the diagonal elements are always reset to the absolute value of their estimates~\cite{myers1976}. This filter is implemented using a UKF without consider analysis, since the tuning of the process noise covariance is expected to account for the uncertainty in the density.
    \item The second filter, denoted as UKF-AC (AC for augmentation and correction), augments the estimation state to compute a correction to a nominal profile as defined in \eqref{eq:ac1} and \eqref{eq:ac2}. The nominal profile used is the least squares exponential fit to the Mars-GRAM training trajectories. This filter is implemented using a UKF without consider analysis as the augmentation of the state mirrors the consider approach, with the distinction that the consider parameter and its covariance are updated. The correction factor is modeled as a random walk, following similar linear adaptation approaches in previous literature~\cite{zegwe2023}. This factor is initialized as the ratio between the true density and the nominal density, with an associated initial covariance of $P_{\mathcal{K},0} = \num{1e-10}$ and process noise of $Q_{\mathcal{K}} = \num{1e-7}$.
    \item The third filter, referred to as USKF-NN (NN for neural network), adheres to the framework detailed in Section~\ref{sec:densityadaptwnn}, where the neural network is trained using the least squares exponential fit to the Mars-GRAM training trajectories. The network is adapted online with the maximum likelihood optimization step, and the filter uses a consider analysis to account for the uncertainty in the density estimate. The consider parameter is modeled as an ECRV centered at one with $\tau = 5$, $P_{ss} = \num{1e-3}$, and an initial covariance of $P_{c,0} = \num{1e-10}$.
\end{itemize}

The parameters used for the maximum likelihood optimization are summarized in Table~\ref{tab:ML-P}. It is important to note that the learning rate within Adam is dynamically lowered with respect to the time step ($k$), as higher gradients are expected when the vehicle reaches maximum dynamic pressure. All filters use the same values for the parameters in the unscented transform, where $\alpha = 1$, $\kappa = 3 - L$, and $\beta = 2$. The initial filter state is sampled from a Gaussian distribution for each Monte Carlo simulation, with the true state as its mean and a diagonal covariance matrix following the uncertainty values as described in Table~\ref{tab:X0-P0}.

\begin{table}[htbp]
    \centering
    \caption{Maximum likelihood parameters used in the USKF-NN}
    \begin{tabular}{lcc}
    \hline \hline
       Parameter  & Value \\ \hline  
       $\beta_1$ & 0.1 \\
       $\beta_2$ & 0.9\\
       $p_{\max}$ & 1 \\
       $\alpha_{\mathcal{L}}$ & $0.01/k$ \\
       $\mathcal{T}$ & 1 \\ \hline \hline
    \end{tabular}
    \label{tab:ML-P}
\end{table}

\subsection{Monte Carlo Results} 

To quantify the accuracy of each filter, the time-averaged root mean squared error (RMSE) is used, defined as:
\begin{equation}
    \mathrm{RMSE}_i  = \frac{1}{N_k}\sum_{k = 1}^{N_k}\frac{1}{N_m}\sum_{j = 1}^{N_m}\sqrt{ \pr{x_i(k)^{(j)} -\hat{x}^+_i(k)^{(j)}}^2},
\end{equation}
where $N_k$ is total number of discrete steps in each simulation, $N_m$ is the number of Monte Carlo simulations, $x_{i}(k)^{(j)}$ is the $i$-th true state and $\hat{x}^{+}_{i}(k)^{(j)}$ is the $i$-th posterior estimate. Table~\ref{tab:ST-RMSE} presents the time-averaged RMSE obtained for each state using the three studied filtering strategies. The USKF-NN consistently outperforms both the UKF-CM and the UKF-AC, demonstrating significantly lower estimation errors across all states. This improvement shows that the newly introduced adaptation scheme can outperform previous adaptive strategies. Notably, the UKF-AC obtains a lower estimation error than the UKF-CM for most of the estimated states, attributed to the adaptive density scheme employed in this filter.

\begin{table}[H]
    \centering
    \caption{Time-averaged root mean squared error for the three filtering strategies}
    \begin{tabular}{lccc}
    \hline \hline
    Parameter  & UKF-CM & UKF-AC & \textbf{USKF-NN} \\ \hline 
    $\tilde{r}$ (m) & \num{3.0631e+03} & \num{4.5090e+03} & \textbf{\num{6.5492e+01}} \\
    $\tilde{\phi}$ (deg) & \num{5.7906e-03} & \num{6.7430e-04} & \textbf{\num{2.3474e-04}} \\ 
    $\tilde{\theta}$ (deg) & \num{1.4322e-01} & \num{1.4548e-02} & \textbf{\num{6.1734e-04}} \\
    $\tilde{v}$ (m/s) & \num{0.7515e+01} & \num{0.8712e+01} & \textbf{\num{5.5504e-01}} \\
    $\tilde{\gamma}$ (deg) & \num{3.7167e+01} & \num{8.3381e-01} & \textbf{\num{1.2342e-02}} \\
    $\tilde{\psi}$ (deg) & \num{1.1331e-02} & \num{1.5826e-03} & \textbf{\num{1.1364e-03}} \\
    $\tilde{B}$ (m$^2$/kg) & \num{8.1736e-05} & \num{2.0957e-04} & \textbf{\num{4.3353e-05}} \\
    $\tilde{L}/D$ (n.d.) & \num{1.8454e-03} & \num{1.7565e-03} & \textbf{\num{1.7378e-03}} \\ \hline \hline
    \end{tabular}
    \label{tab:ST-RMSE}    
\end{table}

\begin{figure*}[h]
    \centering
    \includegraphics[width = \textwidth]{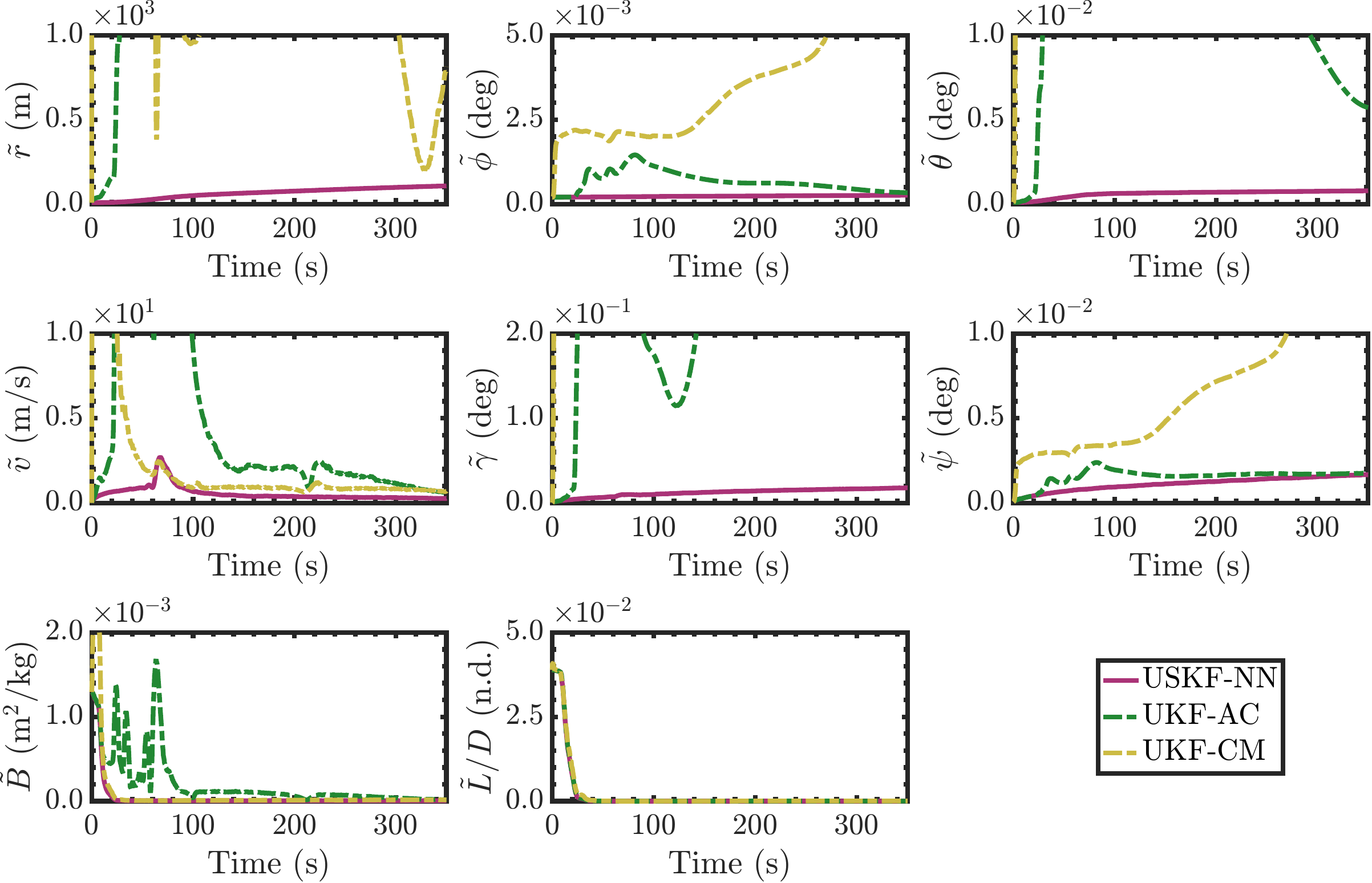}
    \caption{Root mean squared error for each state obtained with each of the three filtering strategies.}
    \label{fig:rmse}
\end{figure*}

Figure~\ref{fig:rmse} shows the root mean squared error as a function of time for each estimation state and each of the three filtering strategies. From this figure, it can be seen that the USKF-NN outperforms every filter for all considered time steps, achieving the lowest estimation error across all states. In contrast, the UKF-AC and the UKF-CM show high estimation errors, particularly in states affected by the atmospheric density. The states not directly dependent on the density for dynamic propagation, such as the heading azimuth (since the bank angle is set to zero), the inverse of the ballistic coefficient, and the lift-to-drag ratio, have similar estimation errors across all the filtering strategies. There is an exception for the heading azimuth in the UKF-CM, where the estimation error quickly diverges at the start of the trajectory. However, it is important to note that the UKF-CM tends to converge faster than the UKF-AC in some of the observable states such as the velocity and inverse of the ballistic coefficient (A more comprehensive observability analysis is detailed in~\cite{levesque2007}). The comparison of these three filters highlights the significant improvement in accuracy achieved when the onboard atmospheric density is adapted using the maximum likelihood optimization in the USKF-NN scheme, especially in states directly influenced by atmospheric conditions during dynamic propagation.

\begin{figure*}[h]
    \centering
    \includegraphics[width = \textwidth]{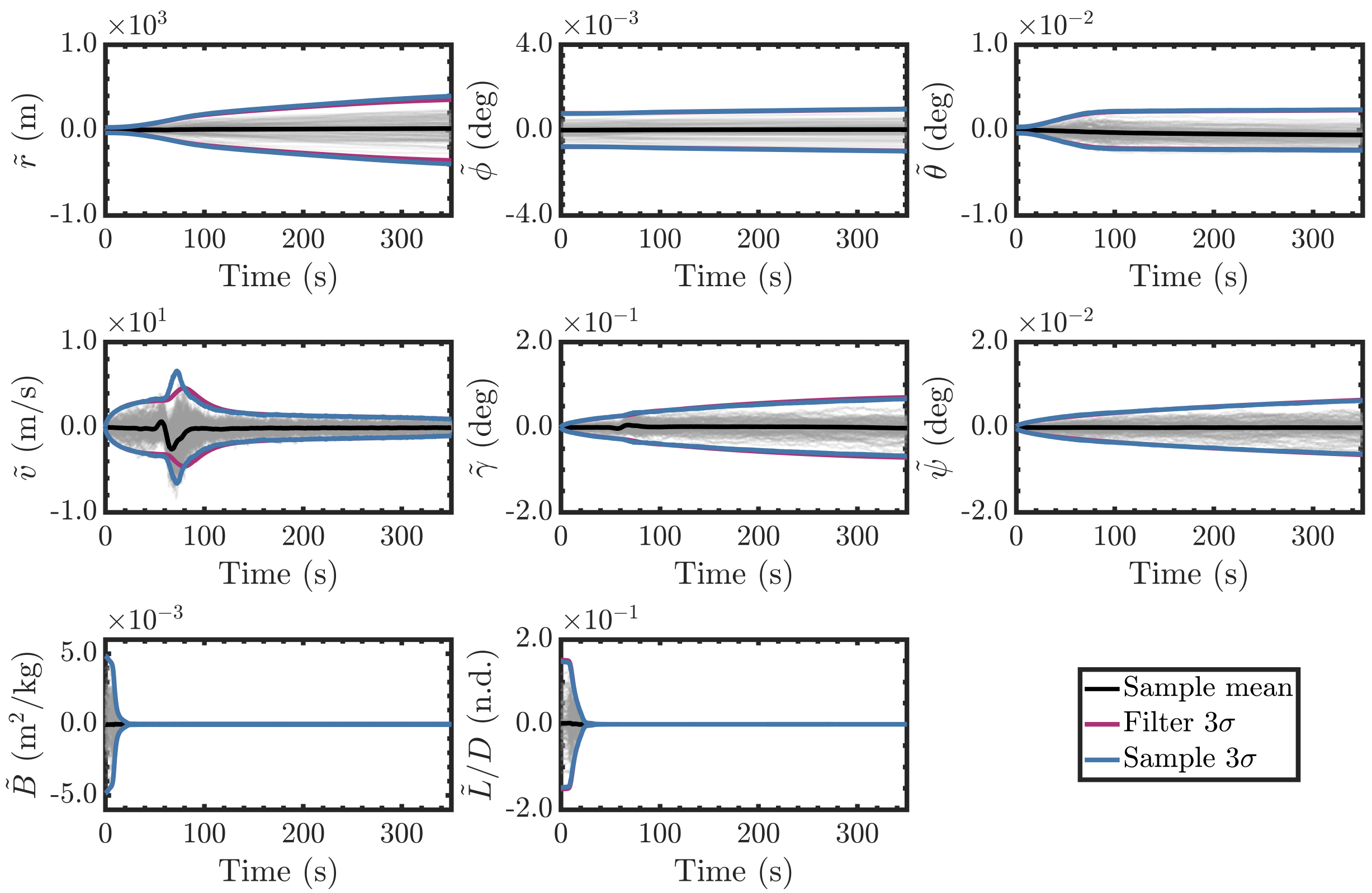}
    \caption{Estimation error as a function of time for the results obtained with the USKF-NN. The gray lines show different Monte Carlo trajectories.}
    \label{fig:una}
\end{figure*}

\begin{figure*}[h]
    \centering
    \includegraphics[width = \textwidth]{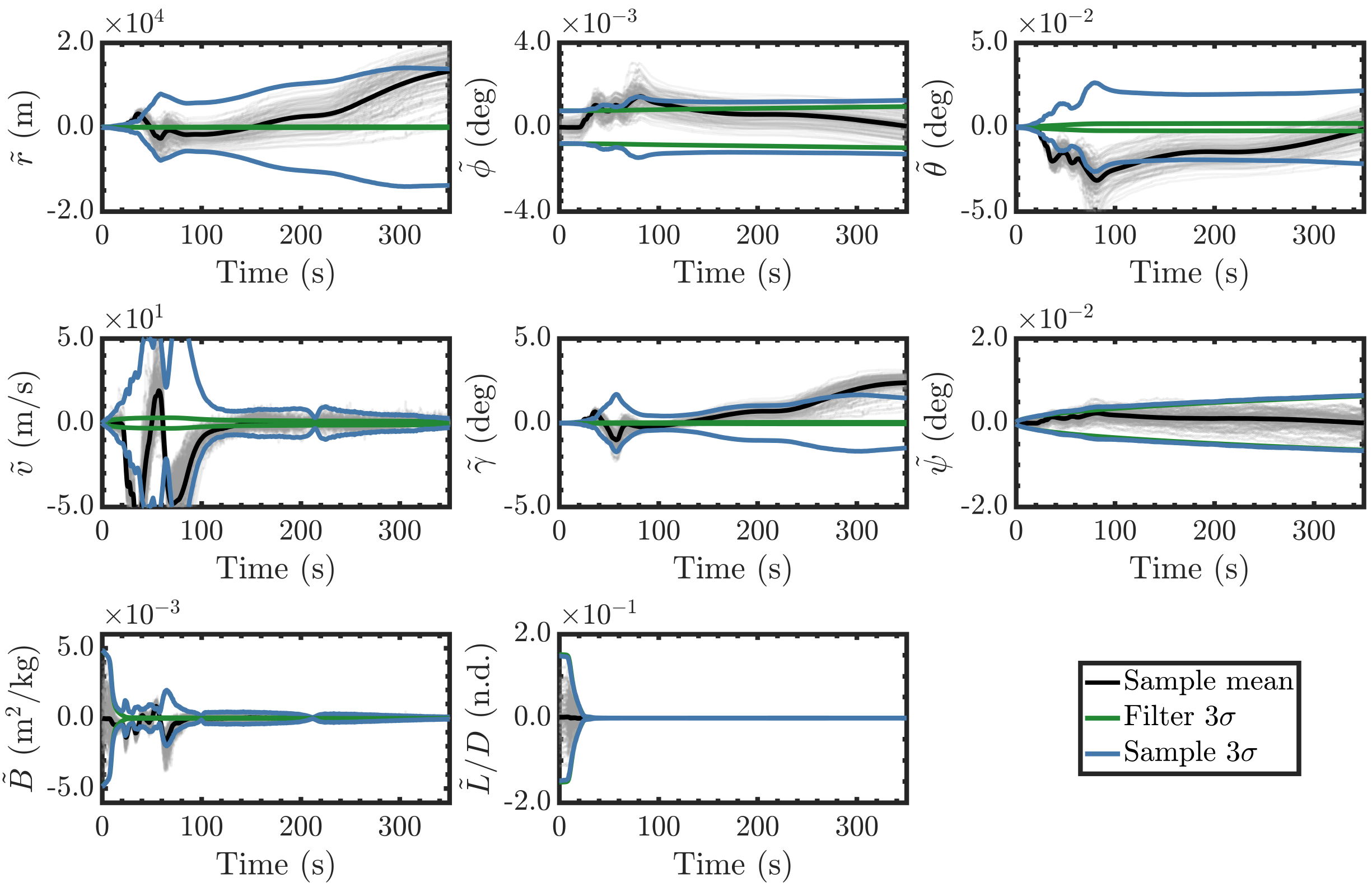}
    \caption{Estimation error as a function of time for the results obtained with the UKF-AC. The gray lines show different Monte Carlo trajectories. Note that, for improved readability, the $\bm{y}$-scale differs from that in Fig.~\ref{fig:una} for variables $\tilde{r}$, $\tilde{\theta}$, $\tilde{v}$, and $\tilde{\gamma}$.}
    \label{fig:uca}
\end{figure*}

The consistency improvement of the USKF-NN over the other filtering strategies can be further analyzed in Fig.~\ref{fig:una} and Fig.~\ref{fig:uca}. These figures show the error plots obtained with the USKF-NN and the UKF-AC, respectively. The error plots for the UKF-CM are not presented, as they were not considered particularly informative for the analysis, as the filter is highly inconsistent. The estimates obtained with the USKF-NN exhibit minimal bias as the estimation error remains close to zero mean for all states. In addition, the predicted covariance of this filter closely aligns with the sample variance of the errors, showing that the USKF-NN is a consistent filter for this scenario. It is important to note that the estimates for the velocity struggle to be zero mean when maximum dynamic pressure is reached (although the covariance of the filter accounts for this error), but the filter quickly recovers maintaining unbiased estimates for the remainder of the flight. Conversely, the UKF-AC struggles to obtain unbiased estimates as the errors are highly biased (the estimates for the heading azimuth remain mostly unbiased since $\sigma = 0$, thus making the dynamic propagation of this state not directly affected by density). The disparity between the true atmospheric density and the onboard estimated density can rapidly introduce bias into the estimates, explaining the sub-optimal performance of the UKF-AC in this regard. This filter also struggles in capturing the uncertainty of its estimates as the predicted filter covariance does not match the variance of the estimation errors.

With the measurement setup used in this work, the estimated states are not entirely observable. Consequently, only the velocity and aerodynamic parameters appear as observable states, as seen in Fig.~\ref{fig:una}. Nevertheless, the USKF-NN achieves consistent results that can potentially be improved as more measurements become available in the descent and landing phases. The results from the UKF-AC are uninformative and it would be difficult to recuperate the estimates once more measurements are available.

Given the comparable strategies employed by the UKF-AC and the USKF-NN, both incorporating online adaptation of the atmospheric density model, it is important to quantify the error in the density estimates associated with this online adaptation. To measure the difference between the true atmospheric density and the estimate generated by each filter, the time-averaged root mean squared percentage error (RMSPE) is used. In this work, this metric is defined as follows:
\begin{equation}
    \mathrm{RMSPE} = \frac{1}{N_k}\sum_{k=1}^{N_k}\frac{1}{N_m}\sum_{j = 1}^{N_m}\sqrt{\pr{\frac{\rho(k)^{(j)} -\hat{\rho}(k)^{(j)}}{\rho(k)^{(j)}}}^2} \cdot 100\%,
\end{equation}
where $N_k$ is total number of discrete steps in each simulation, $N_m$ is the number of Monte Carlo simulations, $\rho(k)^{(j)}$ is the true atmospheric density and $\hat{\rho}(k)^{(j)}$ is the estimated density. Table~\ref{tab:unavsuca} displays the time-averaged RMSPE of the density estimates for both the UKF-AC and USKF-NN. The estimates derived from the USKF-NN significantly outperform those from the UKF-AC, evidenced by the substantial difference in RMSPE. The USKF-NN excels over the UKF-AC by nearly two orders of magnitude, representing a smaller error in relation to the true density.

\begin{table}[h]
    \centering
    \caption{Time-averaged root mean squared percentage error of the estimated densities}
    \begin{tabular}{lcc}
    \hline \hline
       Filter  & RMSPE (\%) \\ \hline 
       UKF-AC & \num{32.4933}\\
       \textbf{USKF-NN} & $\mathbf{\num{0.6960}}$\\ \hline \hline
    \end{tabular}
    \label{tab:unavsuca}
\end{table}

\begin{figure}[htbp]
    \centering
    \includegraphics[width = 0.5\textwidth]{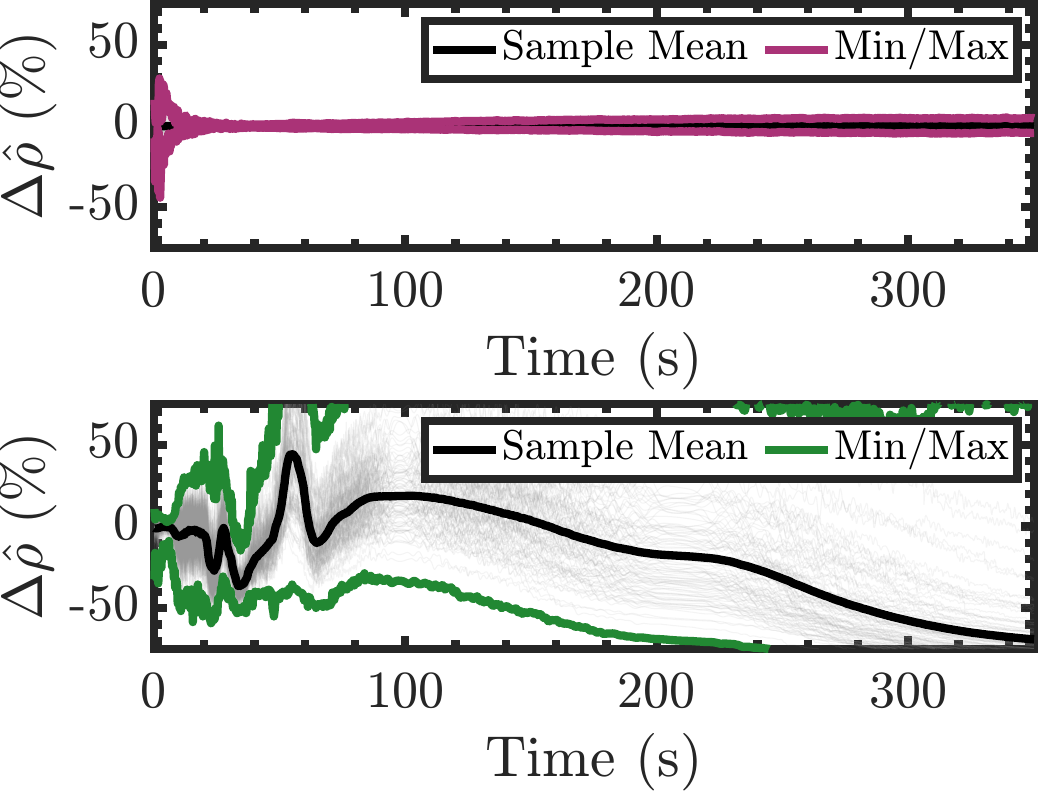}
    \caption{Percentage error of the estimated density as a function of time for the USKF-NN (top) and the UKF-AC (bottom). The gray lines show different Monte Carlo trajectories.}
    \label{fig:unavsuca}
\end{figure}

The difference in density estimates between these two filters can be further analyzed in Fig.~\ref{fig:unavsuca}, which shows the percentage error of the estimated density as a function of time for both solutions. The figure clearly shows the superior performance of the USKF-NN over the UKF-AC. The density estimates generated by the USKF-NN closely track the true density, exhibiting a very low percentage error. In contrast, the UKF-AC struggles to accurately follow the true density, resulting in relative errors that exceed the 50\% mark. As previously mentioned, the USKF-NN adapts the density model in a nonlinear fashion with respect to the measurement, while the UKF-AC uses a linear update to this model. The discernible difference in error between the USKF-NN and the UKF-AC explains the improvement in estimated states, as depicted in Fig.~\ref{fig:una}, highlighting the performance gains achieved through this nonlinear adaptation technique.

\section{Limitations and Future Work} \label{sec:limitations}

While this work has shown the potential benefits of using neural networks as an adaptive estimation method over conventional approaches, such as covariance matching and state augmentation and correction, it is essential to acknowledge certain limitations and assumptions inherent in the methodology.

In this work, the attitude of the vehicle, particularly the bank angle and angle of attack, were assumed to be constant. However, the angle of attack typically varies in entry applications~\cite{dutta2014b} and bank angle control is a common guidance strategy for Martian entry missions~\cite{lu2014}. This highlights the need to consider non-constant attitudes in future work. Nevertheless, given that the methodology described in this work is independent of control inputs, the maximum likelihood adaptation scheme is expected to remain consistent in the presence of bank angle control. Future iterations of this work will address the consideration of non-constant bank angles for the integration of the network into a closed-loop simulation.

Additionally, this work assumes that the vehicle can recover onboard the dynamic pressure and convective heating rate up to an additive white-noise term, rather than processing raw observations. More precise methods, such as Newtonian flow modeling, could be used to extend this work into a higher fidelity testing scenario. In Newtonian flow modeling, surface pressures at each port are determined using approximations based on factors like port location, angle of attack, angle of side-slip, and Mach number~\cite{beck2009}. In this case, special attention should be given to the loss function used in \eqref{eq:minproblem} since the measurement noise may not adhere to a simple additive model.

In the maximum likelihood optimization step, the loss function exclusively uses the current measurement. To account for previous measurements, the optimization step is performed using the Adam optimizer, which retains memory of past gradients without incurring the time complexity increase associated with incorporating all previous measurements. However, including a window ($w_o$) of past and current measurements in the loss function could potentially improve filtering performance. It is worth noting that such adjustments linearly increase the time complexity of the optimization step. In the current framework, this modification would increase the time complexity of the optimization step to $\mathcal{O}(2\cdot h_o \cdot n_o \cdot w_o)$.  Therefore, if implemented, it would be important to find an optimal window length where the accuracy gain outweighs the computational cost.

Finally, an advantage of the proposed method is that the nominal model used to train the neural network offline can be easily changed without needing to modify the online adaptation portion. Therefore, instead of training the neural network using an exponential model, it could be trained on a higher-fidelity model that better represents the expected true density profile. Additionally, this approach is not constrained to using only one neural network. Multiple neural networks could be trained at varying altitudes using a multi-height exponential model. In this way, each neural network is adapted online when the estimated position falls within the range corresponding to that network. These modular capabilities make the presented framework a good candidate for training on higher-fidelity models before actual deployment, which could improve its robustness.

\section{Conclusions} \label{sec:conclusions}

This work introduces a new approach to onboard adaptive filtering for Martian entry using a neural network to estimate atmospheric density, and employing a consider analysis to account for the uncertainty in this parameter. The network is trained offline on a simple exponential atmospheric density model, and its parameters are dynamically adapted in real time to address any disparities between the true and estimated atmospheric densities. The adaptation of the network is formulated as a maximum likelihood problem by leveraging the measurement innovations of the filter to identify the optimal network parameters. The incorporation of a neural network in the current framework enables the use of stochastic optimizers from the machine learning domain, such as Adam, within the context of maximum likelihood adaptive estimation. These stochastic optimizers have been designed for neural networks and are known for their accuracy and efficiency.

Three filtering strategies were implemented and compared to evaluate the effectiveness of the proposed adaptation scheme. An unscented Kalman filter with covariance matching, an unscented Kalman filter with density adaptation by state augmentation, and the newly presented approach were tested through Monte Carlo analysis with different possible atmospheric profiles sampled from Mars-GRAM. The new approach outperformed the other two filters in terms of the estimation error, yielding mostly unbiased and consistent estimates. In addition, the new approach successfully adapted the onboard density model to match the atmospheric profiles provided by Mars-GRAM within an error of 0.7\%.

\section*{Acknowledgment}

This material is based on research sponsored by Air Force Office of Scientific Research (AFOSR) under award number: FA9550-22-1-0419.

\bibliography{sample.bib}

\end{document}